\documentclass[11pt,doublespace,letterpaper]{article}

\usepackage{braket,amsfonts}
\usepackage{cvpr}
\usepackage{times}
\usepackage{amsfonts} 
\usepackage{epsfig}
\usepackage{graphicx}
\usepackage{amssymb}

\cvprfinalcopy 

\PassOptionsToPackage{numbers, compress}{natbib}

\usepackage[utf8]{inputenc} 
\usepackage[T1]{fontenc}    
\usepackage{hyperref}       
\usepackage{url}            
\usepackage{booktabs}       
\usepackage{amsfonts}       
\usepackage{nicefrac}       
\usepackage{microtype}      
\usepackage{graphicx}

\title{Penalized k-means algorithms for finding the correct 
number of clusters\\ in a dataset}

\author{
Behzad Kamgar-Parsi$^{\star}$ \hspace*{1.0cm} Behrooz Kamgar-Parsi$^{\dagger}$\\
\\
$^{\star}$Office of Naval Research, Arlington, VA 22203\\
{\small{\tt behzad.kamgarparsi@navy.mil}}\\
$^{\dagger}$Naval Research Laboratory, Washington, DC 20375\\
{\small{\tt behrooz.kamgarparsi@nrl.navy.mil}}\\
}

\begin{document}

\maketitle

\begin{abstract}
In many applications we want to find the number of clusters in
a dataset. A common approach is to use the penalized k-means algorithm with an
additive penalty term linear in the number of clusters. 
An open problem is estimating the value of the coefficient of the penalty term.
Since estimating the value of the coefficient in a 
principled manner appears to be
intractable for general clusters, we investigate {\em ideal
clusters}, i.e.  identical spherical clusters with no 
overlaps and no outlier background noise. 
In this paper: (a) We derive, for the case of ideal clusters, rigorous 
bounds for the coeffcient of the additive penalty.  
Unsurprisingly, the bounds depend on the correct
number of clusters, which we want to find in the first place. 
We further show that additive penalty, even for this simplest case of
ideal clusters, typically produces a weak and often
ambiguous signature for the correct number of clusters. (b) As an alternative,
we examine the k-means with multiplicative penalty, and show that
this parameter-free formulation has a stronger, and less often ambiguous,
signature for the correct
number of clusters. We also empirically investigate certain types of
deviations from ideal 
cluster assumption and show that combination of k-means with additive 
and multiplicative penalties can resolve ambiguous solutions. 

\end{abstract}

\section{Introduction}
Finding the correct number of clusters in a dataset is an open,
 important problem in many applications. 
A widely used algorithm for this purpose is k-means, and while
it is appropriate only for same-size spherically symmetric
clsuters, in practice it is often used beyond its valid range also
for clusters with vastly different shapes and sizes.
Because of its applicability in many areas including
image and pattern analysis, 
statistics, information theroy, data reduction and others, there are  
a number of excellent treatments of k-means in the literature. Here we cite 
only a few \cite{Blum-18}\cite{Duda-03}\cite{Gersho-92}\cite{Jain-88}\cite{Linde-80}\cite{Manning-08}\cite{Murphy-12}\cite{Vidal-16}.
For the sake of completeness, we include a brief
description of the k-means here as well.

The problem may be stated as: 
Given a set of $N$ unlabeled data points $\{\mathbf{x}_i\}$, $i=1,\ldots, N$, 
in a $d$-dimensional feature space, we want to partition the data 
into $k$ clusters $\{{\cal C}_j\}$, $j=1,\ldots,k$,
such that the empirical error $E$ (or cost function,  
distortion, sum of within-cluster variances),
\begin{equation}
E=\sum_{j=1}^{k}\,\sum_{\mathbf{x}_i\in\,{\cal C}_j}\,\parallel \mathbf{x}_i - 
\mathbf{c}_j\parallel^2,
\,\,\,\,\,\mbox{with}\,\,\,\,\,
\mathbf{c}_j=\frac{1}{N_j}\,\sum_{\mathbf{x}_i\in\,{\cal C}_j}\,\mathbf{x}_i,
\end{equation}
is minimized. Here $\mathbf{c}_j$ denotes the centroid of cluster ${\cal C}_j$,
$N_j$ is the number of data points that are in cluster ${\cal C}_j$, and
$N=\sum_{j=1}^{k}N_j$. The k-means algorithm is a popular method for
solving this non-convex optimization problem. 

In the k-means algorithm we start by randomly selecting $k$ cluster centroids, 
and then use the Lloyd iterations to obtain a solution.  The Lloyd
iteration~\cite{Lloyd-82} is a simple Expectation-Maximization (EM) process,
more precisely a hard EM process~\cite{Murphy-12},
where the two steps are: (i) assign the points that are closest to 
a cluster centroid to that centroid, (ii) recompute the cluster centroids.
Then repeat the two steps until cluster memberships no longer 
change. This process monotonically decreases the error and typically converges
to a local minimum. In order to potentially find a
better solution, the algorithm is restarted many times with different 
initial cluster centroids and then the clustering with the smallest error
is taken as {\em the solution}. Various initialization procedures have been
introduced in the literature among which we cite the 
k-means++~\cite{Arthur-07}, and its 
modifications~\cite{Bachem-16}\cite{Baldassi-19},
that generally yield better solutions. 
Also a number of alternative algorithms have been proposed in the 
literature for more efficient solutions. Here we cite only a few:
\cite{Kanungo-02} computes distances more efficiently;
\cite{Li-17} presents a parallel method; \cite{Rose-92}
proposes successive partitioning of data; and  \cite{Kamgar-90}\cite{Kamgar-92} 
proposes a soft k-means algorithm based on the Hopfield network
that generally leads to better solutions, albeit at higher comuptational 
cost. 

We note that these algorithms, even for the simple case of {\em ideal}
clusters (defined below), cannot be proven to find the {\em best
solution}, or yield objective information about the {\em correct} number of 
clusters.  Neither are the results reproducible, because of the random 
selection of initial clusters.

In many applications we do not know the number of clusters a priori, rather
 we want to find the correct number of clusters.
Examples include finding the number of speakers in a
crowded room from audio recordings, the number of vessels in
a high-traffic channel from navigation radar signals, the number of
distinct textures in an image, gene expressions, and
many others. 
Since increasing $k$ leads to solutions with ever
decreasing $E$, the error (1.1) cannot indicate what the correct $k$ may be. 
Therefore, we must use a {\em penalized error} that has
its global minimum at the correct value of $k$.
Penalized error may be formulated with either {\em additive} or 
{\em multiplicative} penalties respectively given by:
\begin{equation}
E^{(a)}=\sum_{j=1}^{k}\,\sum_{\mathbf{x}_i\in\,{\cal C}_j}\,\parallel \mathbf{x}_i - 
\mathbf{c}_j\parallel^2+\lambda f(k),
\end{equation}
\begin{equation}
E^{(m)}=f(k)\,\sum_{j=1}^{k}\,\sum_{\mathbf{x}_i\in{\cal C}_j}\,
\parallel \mathbf{x}_i - \mathbf{c}_j\parallel^2,
\end{equation}
where $f(k)$ is a monotonically increasing function of $k$. 
The common practice is to use the additive 
penalized error $E^{(a)}$. It will have a global
minimum at some finite $k$, which depends on the value of the penalty
coefficient $\lambda$. The problem then becomes the more complicated
simultaneous determination of
the optimal $k$, as well as the optimal set of cluster centroids 
$\{\mathbf{c}_j\}$, $j=1,\ldots,k$. The correct number of clusters depends
critically on the value of coefficient $\lambda$. Obviously, small $\lambda$ 
favors a larger number of clusters, and vice versa, large $\lambda$ leads
to a smaller number of clusters. 

In this paper, we present a rigorous analysis of the penalized k-means with
additive penalty to obtain principled bounds for the value of $\lambda$.
Since treating general
clusters appears to be intractable, we consider ideal clusters that
are the simplest form of clusters. We will describe and justify the rather
strong assumption for using ideal clusters in the next section. But briefly,
ideal clusters are $d$-dimensional spheres, with identical sizes, with no
overlapps, and no outlier background data points.

\subsection*{Related work}

Because finding the correct number of clusters plays an important role in
many applications, 
a large number of methods have been proposed in the literature. 
While many methods have been proposed for the standard non-penalized k-means
formulation (1.1), there is little work on the penalized k-means formulations
(1.2) and (1.3).

For the non-penalized k-means there are two broad approaches, which
maybe called {\em slope-based} and {\em validation-based}. Here we cite 
only a few
relevant papers and the references therein. The slope-based 
approach essentially tries to find a sudden change in the slope of
the clustering error $E$ as a function of $k$, and includes the {\em elbow} 
method, the more elaborate  
 {\em silhouette} method \cite{Rousseeuw-87}, and the related {\em slope 
statistics} method \cite{Fujita-14}. These methods are subjective and
in general unreliable. In the validation-based approach, the data is
divided into a training set and a validation set
\cite{Amorim-16}\cite{Fu-17}. This is a more reliable approach, but the outcome
can be sensitive to how the data is divided into training and validation
sets.

For the penalized k-means there is very little work.
Recently \cite{Kulis-12} derives the penalized
k-means with additive penalty from a Bayesian nonparametric viewpoint.
However, coefficient $\lambda$ is selected as
the largest disatnce any data point can have from
its centroid. Althogh this scheme limits the growth of
number of clusters, it does not find the correct number of clusters.
iTo our knowledge, the multiplicative penalized error $E^{(m)}$ has not been 
studied in the literature. We show that this parameter-free formulation
has properties that appear to be more useful than the additive penalized 
error. We note that the penalized k-means is in the broad class of
slope-based methods, but without being subjective in principle.

In the following sections, we define ideal clusters, analyze penalized 
k-means with additive
and multiplicative penalties, and present a number of selected
experiments that show the validity of theoretical analyses and
certain limits of ideal cluster assumption. 

\section{Ideal Clusters}

Here we define ideal clusters and argue that they are consistent with the
underlying assumptions for k-means clusters even though more restrictive. 
The original k-means clustering 
may be interpreted as: we want to simultaneously estimate
the value of $k$ entities $\{\mathbf{c}_j\}$, from $N$ unlalbeled measurements  
$\{\mathbf{x}_i\}$, and that these measurements have identical independent
normal distributions ${\cal{N}}({\mathbf{c}_j}, \sigma^2 I)$, 
with the same covariance $\sigma^2$, around their 
true values. To do so, we must cluster the measurements into correct
groups and maximize the likelihood, 
\begin{equation}
P=\frac{1}{Z}\prod_{j=1}^{k}\,\prod_{\mathbf{x}_i\in\,{\cal C}_j}\,
e^{\displaystyle{-{\frac{\parallel \mathbf{x}_i - \mathbf{c}_j\parallel^2}{2\sigma^2}}}},
\end{equation}
or equivalently minimize its logarithm $E$ given by (1.1). Here
$Z=(2\pi\sigma^2)^{\frac{dN}{2}}$ is the normalization factor, and $d$ is
the dimension of the feature space.
The underlying assumptions are: (a) clusters are spherically symmetric 
 since the variances are isotropic in all the
Euclidean space dimensions, (b) clusters have the same size since variances 
are identical, (c) clusters are sufficiently
separated so no two clusters overlap in such a way that their
peaks cannot be distinguished from each other, and (d) there are no outlier
data points not belonging to any of the clusters.

We define {\em ideal clusters} to have the following properties: 
(a) clusters are spherical; (b)
have the same size and volume $V$; (c) are non-overlapping, a stronger 
assumption than the original; and (d) have no outliers. We make an additional
assumption (e) that clusters are dense, that is, the data points nearly 
fill the volume of the spheres. This is primarily for computational 
convenience to replace sums with integrals, and implies the
number of data points in each cluster may be approximated by the volume $V$
of the sphere and that $N_j$ is equivalent to $V$.

In regular k-means we know the number of clusters a priori. In penalized 
k-means, however, $k$ is a variable to be determined along with the set of 
cluster centroids $\{\mathbf{c}_j\}$.
The optimization procedure is to vary $k$ and
then for each $k$ find the optimal set $\{\mathbf{c}_j\}$ using the
regular k-means clustering. The expectation is that $E^{(a)}$ as a function 
of the number of clusters behaves as shown in Fig.~1, where the correct 
$k\!=\!K$ is the value at which penalized error is minimum.
This happens {\em if} the regular k-means algorithm finds the optimal set 
$\{\mathbf{c}_j\}$ for each $k$. Since the regular k-means algorithm
cannot guarantee optimality, we restrict the discussion to ideal clusters for 
which we can design an algorithm that is provably optimal.

\begin{figure}[t]
\center{
{\includegraphics[width=1.8in] {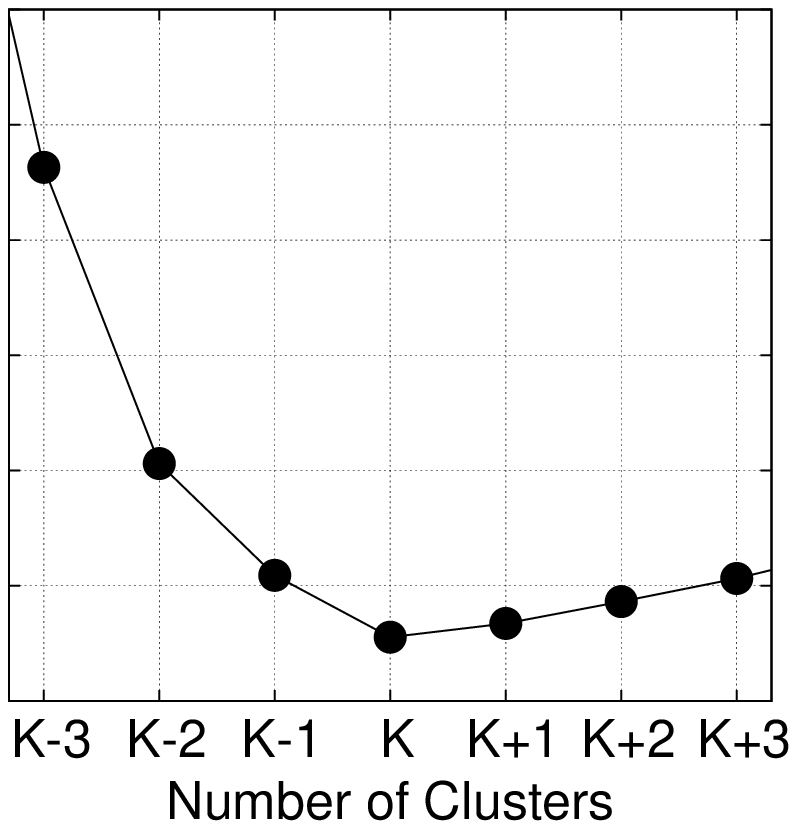}}
\hspace*{-0.4cm}
{\includegraphics[width=1.8in] {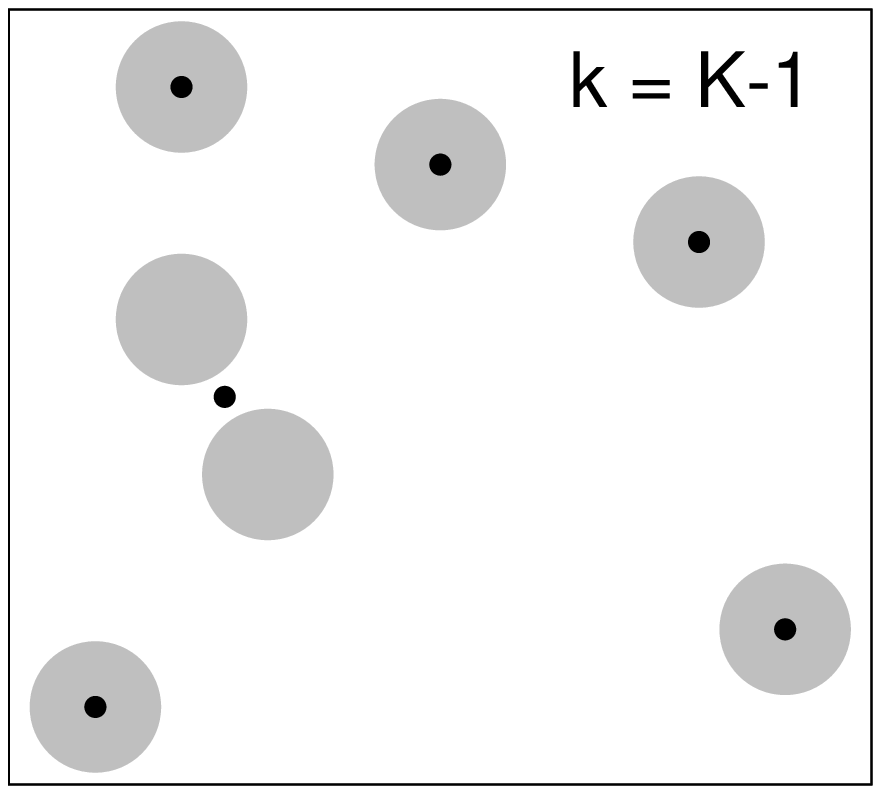}}
\hspace*{-0.6cm}
{\includegraphics[width=1.8in] {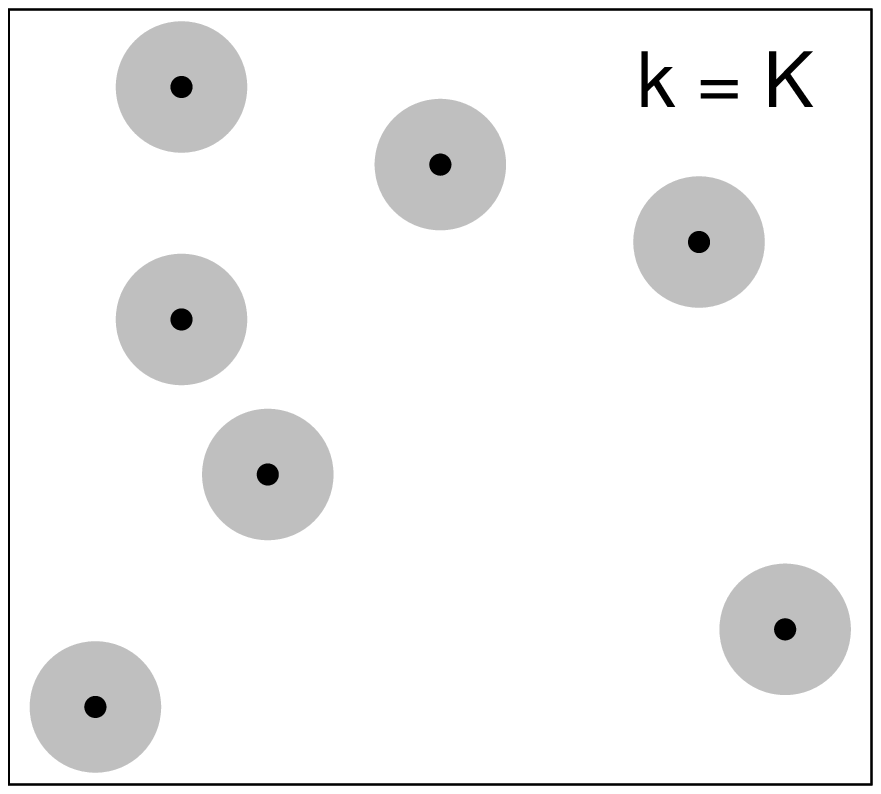}}
\hspace*{-0.6cm}
{\includegraphics[width=1.8in] {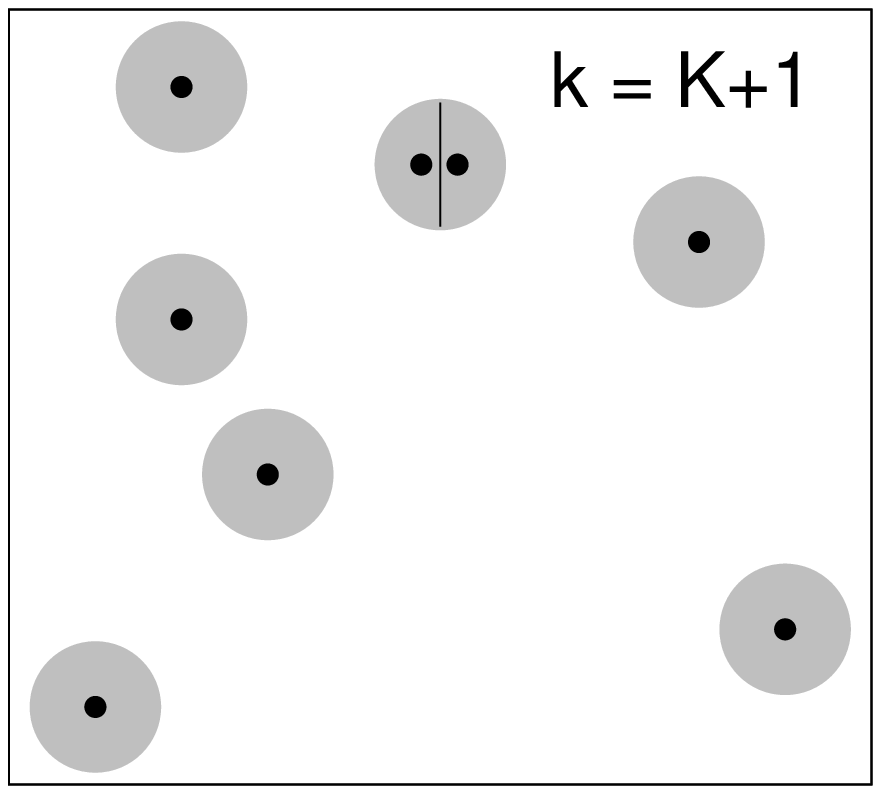}}
}
\caption{({\bf Left}) Sketch of a penalized error with its global minimum
at the correct number of clusters $K$.  ({\bf Right}) A 2D           
example with $K\!=\!7$ ideal clusters, illustrating
results of the optimal algorithm for $k\!=\!K-1$ with 5 spheres and 1   
dumbbell, for $k\!=\!K$ with 7 spheres, and for $k\!=\!K+1$ with 6 spheres and
2 half-spheres.}
\end{figure}

\subsection{Optimal Clustering Algorithm}

The algorithm that is provably optimal for ideal clusters is as follows. 
\begin{itemize}
\item
{Step 1:}
Choose a data point and save it as the initial centroid for the 
first cluster $\mathbf{c}_{1}^{0}$. (The algorithm is insensitive
to the initial choice; we typically choose the point closest
to the origin of the feature space.)
\item
{Step 2:} Choose the data point farthest
from $\mathbf{c}_{1}^{0}$. Save it as the initial centroid for the
second cluster $\mathbf{c}_{2}^{0}$. Using these initial centroids, run the
Lloyd iteration until convergence to $\mathbf{c}_{j}$, $j=1,2$. 
\item
{Step 3:} Choose 
the data point farthest from both $\mathbf{c}_{1}^{0}$ and 
$\mathbf{c}_{2}^{0}$ Save it as the initial centroid for the third 
cluster $\mathbf{c}_{3}^{0}$. Using these initial centroids, run the Lloyd
iteration until convergence to  $\mathbf{c}_{j}$, $j=1,2,3$. 
\item
{Step 4 and higher:} At step $M$ continue 
choosing the farthest point to all the previous $M-1$ initial centroids 
and save it as the initial centroid for the $M$-th cluster. Using
these initial centroids, run the Lloyd iteration until convergence
to $\mathbf{c}_{j}$, $j=1,\ldots,M$.  
\end{itemize}
We note that this initialization
procedure may be considered as a deterministic version of 
k-means++ initialization~\cite{Arthur-07}.

When $k\!=\!K$ ($K$ being the correct number of clusters), the clustering 
is optimal. 
The proof is straightforward, because each initial centroid is in a distinct 
cluster, and since clusters are non-overlapping each initial centroid converges 
to the true centroid of the cluster to  which it belongs. 
When $k\!=\!K+1$ one of the clusters splits into two equal half-spheres, 
since that 
cluster contains two of the initial clusters. And when $k\!=\!K-1$ two adjacent 
clusters merge and form a dumbbell cluster. Fig.~1 illustrates these cases
for an example with $K\!=\!7$.

\subsection{Errors for Individual Clusters}

When the number of clusters is close to the correct value, 
individual ideal clusters take on three different shapes, namely, sphere 
with clustering error $E_s$, half-sphere with error $E_h$, and dumbbell 
with error $E_d$. These individual clustering errors are given by,


\begin{equation}
\left\{
\begin{array}{ll}
E_s&=VR^2\alpha\\
 & \\
E_h&=VR^2\beta\\
 &  \\
E_d&=2E_s+2VL^2=2(VR^2\alpha+VL^2)
\end{array}
\right.
\end{equation}
where $V$ is the volume of the $d$-dimensional sphere with radius $R$,
\begin{equation}
V=\frac{\pi^{\frac{d}{2}}}{\Gamma(\frac{d+2}{2})}R^d,
\end{equation}
$L$ is the distance of the center of the dumbbell to the center of either of
its spheres,  
\begin{equation}
\alpha=\frac{d}{d+2},
\,\,\,\,\,
\beta=\frac{1}{2}\,(\alpha - \gamma^2),\,\,\,\,\,  
\gamma=\frac{\Gamma(\frac{d+2}{2})}{\sqrt{\pi}\,\Gamma(\frac{d+3}{2})},
\end{equation}
and $\Gamma(\cdot)$ is the generalized factorial function. Also
$\gamma=\rho/R$, where $\rho$ is the distance of the half-sphere centroid 
from the equator plane.
We refer to Appendix A for derivations of (2.5) and (2.6). 
Also see \cite{Blum-18} for aspects of statistics in high-dimensional spaces. 

\section{Additive Penalty}

Here we analyze the linear additive penalty $f(k)=k$ to obtain the bounds
for the penalty coefficient $\lambda$. 
Other functional forms of penalty are
discussed in more detail in Appendix B and mentioned briefly at the end of
this section. The conclusions, however, remain the same. 

Let $K$ denote the correct number of spherical (ideal) clusters. 
The case with $K\!-\! 1$ clusters then has $K\!-\!2$ spheres and 
one dumbbell,
and the case with $K\!+\!1$ clusters has $K\!-\!1$ spheres and 2 half-spheres.
Thus, the clustering errors for the non-penalized case $E$ in (1.1) are given 
by:
\begin{equation}
\left\{
\begin{array}{ll}
E_{K-1}&=(K-2)E_s+E_d=K\,VR^2\alpha+2VL^2\\
 & \\
E_K&=KE_s=K\,VR^2\alpha\\
 &  \\
E_{K+1}&=(K-1)E_s+2E_h=(K-1)VR^2\alpha+2VR^2\beta.
\end{array}
\right.
\end{equation}

The clustering error obviously decreases as we increase the number of
clusters, and it is straight forward to show this for the ideal case:
\begin{equation}
\left\{
\begin{array}{lll}
\delta_{K-1,K}\!\!&=E_{K-1}-E_{K}=2VL^2>0& \mbox{for all}\,\,\, d\ge 1
\,\,\,\mbox{and}\,\,\,K> 1,\\
 & & \\
\delta_{K,K+1}\!\!&=E_K-E_{K+1}=VR^2(\alpha-2\beta)>0&
\mbox{for all}\,\,\, d\ge 1 \,\,\,\mbox{and}\,\,\,K\ge 1.\\
\end{array}
\right.
\end{equation}
The second inequality is satisfied since $\alpha/2\beta>1$, as may also be
seen in Fig.~8.

For the clustering error $E^{(a)}$ to have the minimum at $K$ we must require:
\begin{equation}
\left\{
\begin{array}{lll}
\Delta_{K-1,K}^{(a)}\!\!&=E_{K-1}-E_{K}=2VL^2-\lambda>0& \mbox{for all}\,\,\, 
d\ge 1
\,\,\,\mbox{and}\,\,\,K> 1,\\
 & & \\
\Delta_{K,K+1}^{(a)}\!\!&=E_K-E_{K+1}=VR^2(\alpha-2\beta)-\lambda<0&
\mbox{for all}\,\,\, d\ge 1 \,\,\,\mbox{and}\,\,\,K\ge 1.\\
\end{array}
\right.
\end{equation}
SInce $R^2(\alpha-2\beta)\!=\!R^2\gamma^2\!=\!\rho^2$, 
and volume $V$ stands for the average number of points in each cluster,
i.e. $V\approx N/K$, the two inequalities may be combined and written as:
\begin{equation}
\frac{N\rho^{2}}{K} < \lambda < \frac{2NL^2}{K}.
\end{equation}
Note that $\lambda$ depends on the correct number of clusters $K$, 
which we want to determine. This introduces
another layer of complexity in using k-means with additive penalty. To
overcome this problem, we propose the following procedure.

\subsection{Procedure for Using Additive Penalty}

\begin{itemize}
\item Begin by assuming $K\!=\!2$. Select a value for $\lambda$ in 
the range given by (3.11). Run the k-means algorithm for $k\!=\!2,\ldots M$, 
where $M$ is very large. If $E^{(a)}$ has a minimum at $k\!=\!2$, then
the assumed $K\!=\! 2$ is a potential solution.
\item  
Next assume $K\!=\!3$. Select a value for $\lambda$ in the corresponding range.
Run the k-means algorithm for $k\!=\!2,\ldots M$.
If $E^{(a)}$ has a minimum at $k\!=\!3$, then the assumed
$K\!=\! 3$ is a potential solution.
\item 
Then assume $K\!=\!4$, and repeat the process. If $E^{(a)}$ has a minimum 
at $k\!=\!4$, then $K\!=\! 4$ is a potential solution. 
\item And so on assuming $K\!=\!5, 6, \ldots$ up to a very large value.
\end{itemize}

To make this procedure consistent and repeatable, we propose to select
the midpoint of the range given in (3.11) as the value for $\lambda$. That is,
\begin{equation}
\lambda = \frac{N(\rho^2+2L^2)}{2K} \approx \frac{NL^2}{K}.
\end{equation}
The approximation is reasonable since (a) $\rho\!=\!R\gamma$ and 
$\gamma^2\!=\!0.18$ 
for $d\!=\!2$ and goes to zero as the 
dimension $d\rightarrow\infty$, and (b) the $\rho^2$ term may be neglected
since $L\ge R$ for non-overlapping clusters. 
We use the approximate value for $\lambda$ in the experiments we present later. 
This approximation becomes increasingly better as dimension $d$ increases.
For $L$ we choose the smallest inter-centroid distance, which also depends 
on $K$.

We also considered other penalty functions, namely, $f(k)=\ln k, k^p, e^k$. 
Similar to the linear penalty (3.12), $\lambda$ for these cases also depends on
$K$ and is, respectively (see Appendix B), 
\begin{equation}
\lambda\propto NL^2(1+\frac{1}{K}),\,\,\,\,\,\, 
\lambda\propto \frac{NL^2}{K^{p}},\,\,\,\,\,\,
\lambda  \propto \frac{NL^2}{e^K}.
\end{equation}

\section{Multiplicative Penalty}

In this section, we analyze $E^{(m)}$ in (1.3) with 
linear multiplicative penalty $f(k)=k$, and show that for ideal clusters
it has a natural miniumum at the correct number of clusters 
$K$. To do so, we
show that for clusters with multiplicative penalty $\Delta_{K-1,K}^{(m)}>0$
while $\Delta_{K,K+1}^{(m)}<0$: 
\begin{equation}
\left\{
\begin{array}{lll}
\!\!\!\Delta_{K-1,K}^{(m)}\!\!\!\!&=(K-1)E_{K-1}-KE_K=2(K-1)VL^2-KVR^2\alpha>0&
 \mbox{for}\,\,\, d\!\ge\! 1  
\,\,\,\mbox{and}\,\,\,K\!\ge\! 2,\\
 & & \\
\!\!\!\Delta_{K,K+1}^{(m)}\!\!\!\!&=KE_K-(K+1)E_{K+1}=VR^2[\alpha-2(K+1)\beta]<0&
\mbox{for}\,\,\, d\!\ge\! 2\,\,\,\mbox{and}\,\,\,K\!\ge\! 2.
\end{array}
\right.
\end{equation}
The first inequality is satisfied since $L\!\ge\! R$ and $\alpha\!<\! 1$.
The second inequality holds because $1\!<\!\alpha/2\beta\!<\!2\,$ for
$d\!\ge\! 2$ (see Fig.~8 ).
For the uninteresting $d\!=\!1$ case (clustering on a straight line), for
which $\alpha/2\beta\!=\!4$, the inequality is satisfied when
$K\!\ge\! 3$.

Note that the inequalities happen naturally; unlike the additive penalty
case where we must impose inequalities to ensure $E^{(a)}$ becomes
minimum at $K$.

\section{Experiments}

In this section we present several experiments for selected dimensions and
number of clusters, and also varying degrees of cluster overlaps.
They show that multiplicative penalty has a stronger signature
than additive penalty, and is generally more stable when
the ideal non-overlap assumption is violated. 

To do the experiments we first use the clustering
algorithm presented in Sec. 2.1. Then  for the additive penalty we 
use the procedure presented in Sec. 3.1, which yields 
candidate solutions when the {\em Assumed}
and {\em Estimated} number of clusters are the same.
We also compute the multiplicative penalized error $E^{(m)}$ as a
function of $k$ from (1.3). Its minima are candidate solutions. In cases where
there are ambiguities as to the correct number of clusters $K$, the agreement
between solutions of additive and multiplicative penalties resolves
the ambiguity.

Figures 2 and 3 are both 2D cases with $K\!=\! 10$ clusters each 
with approximately 100 point points. It can be seen that the additive penalty
yields several candidate solutions, while multiplicative penalty yields only one
solution at the correct number of clusters. It can also be seen that the
additive penalized errors are significantly shallower than the multiplicative
penalized error. Also note that as clusters overlap more, as in Fig.~3, both
additive and multiplicative errors become shallower.

\begin{figure}[t]
\center{
{\includegraphics[width=2.0in] {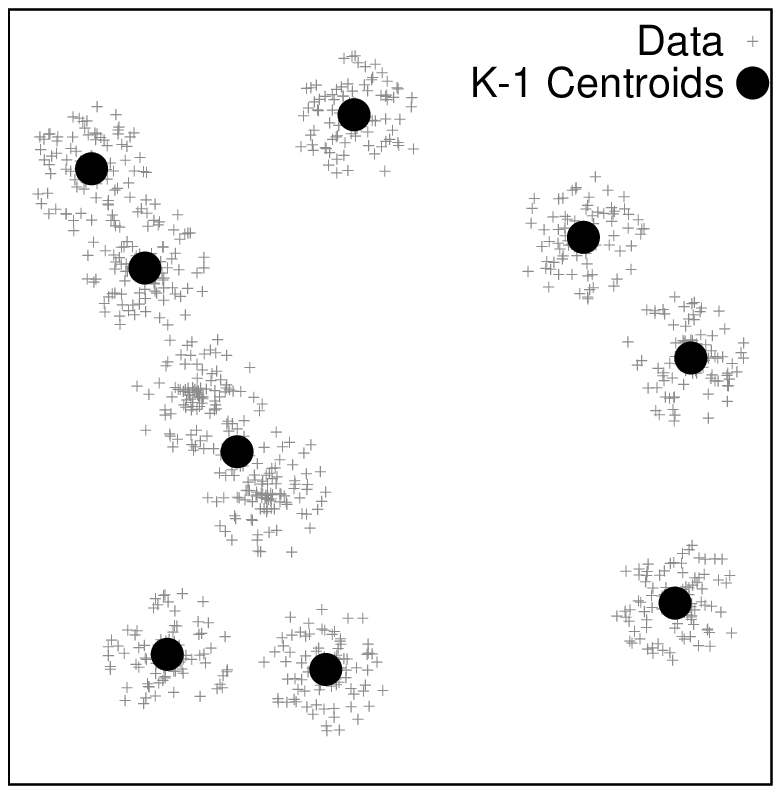}}
\hspace*{-0.5cm}
{\includegraphics[width=2.0in] {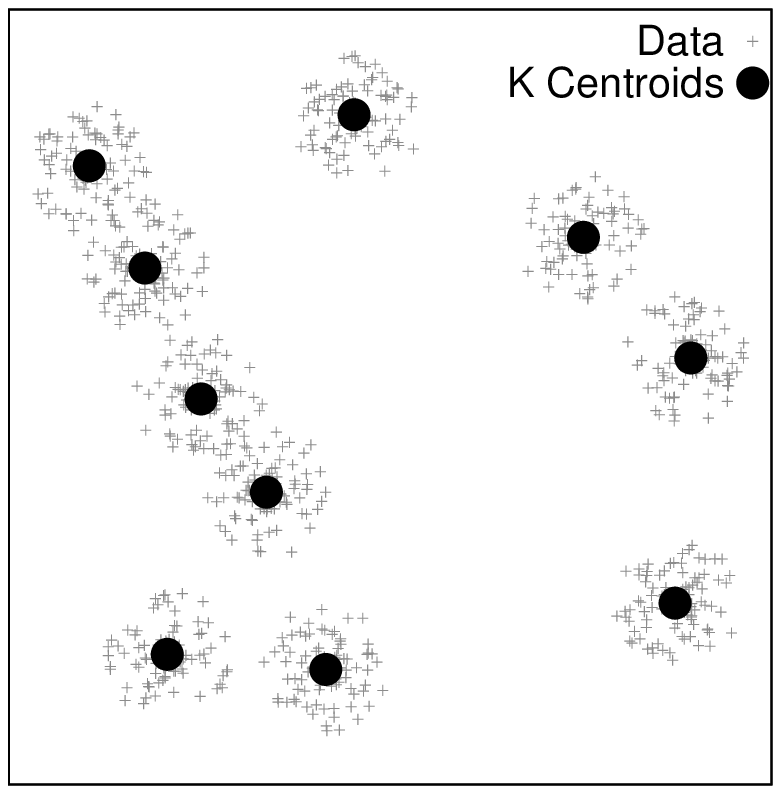}}
\hspace*{-0.5cm}
{\includegraphics[width=2.0in] {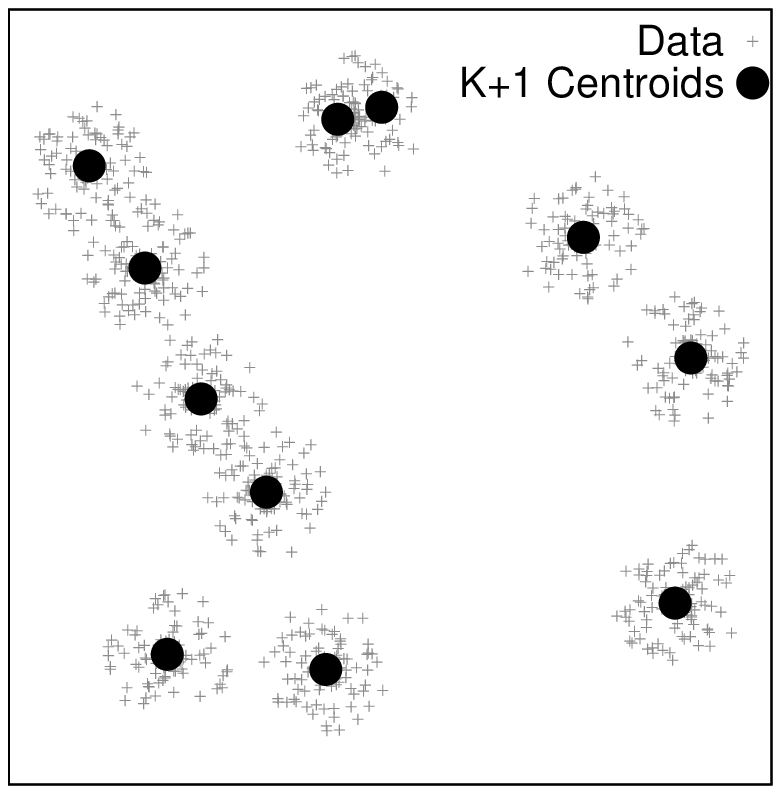}}\\
{\includegraphics[width=2.0in] {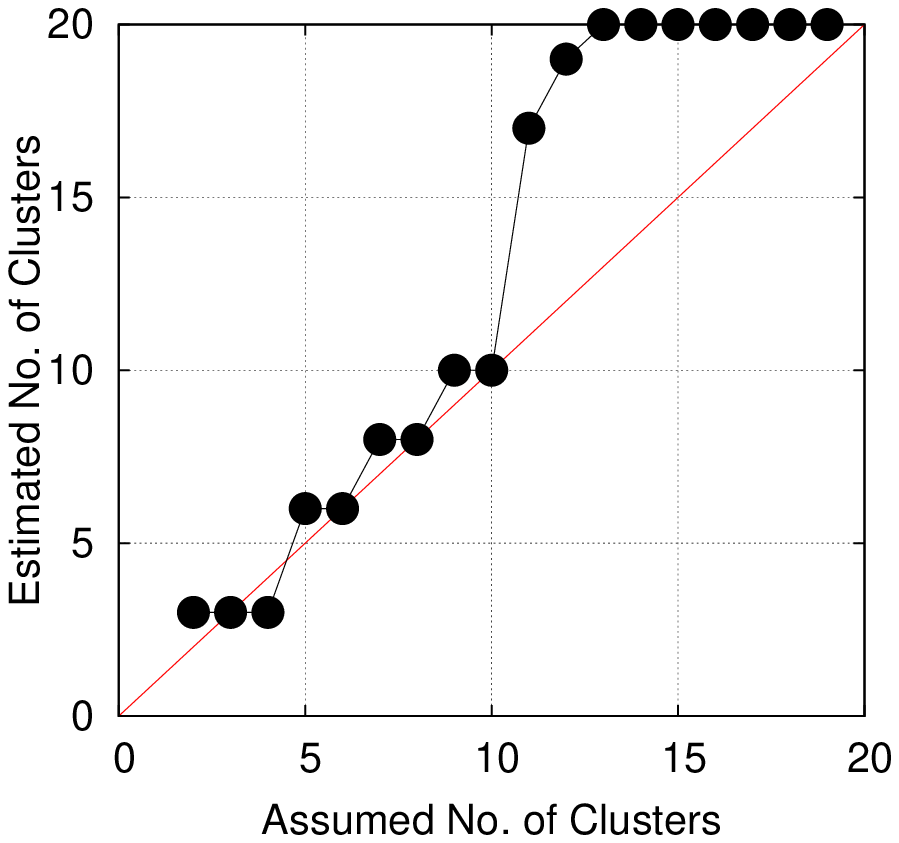}}
\hspace*{-0.6cm}
{\includegraphics[width=2.0in] {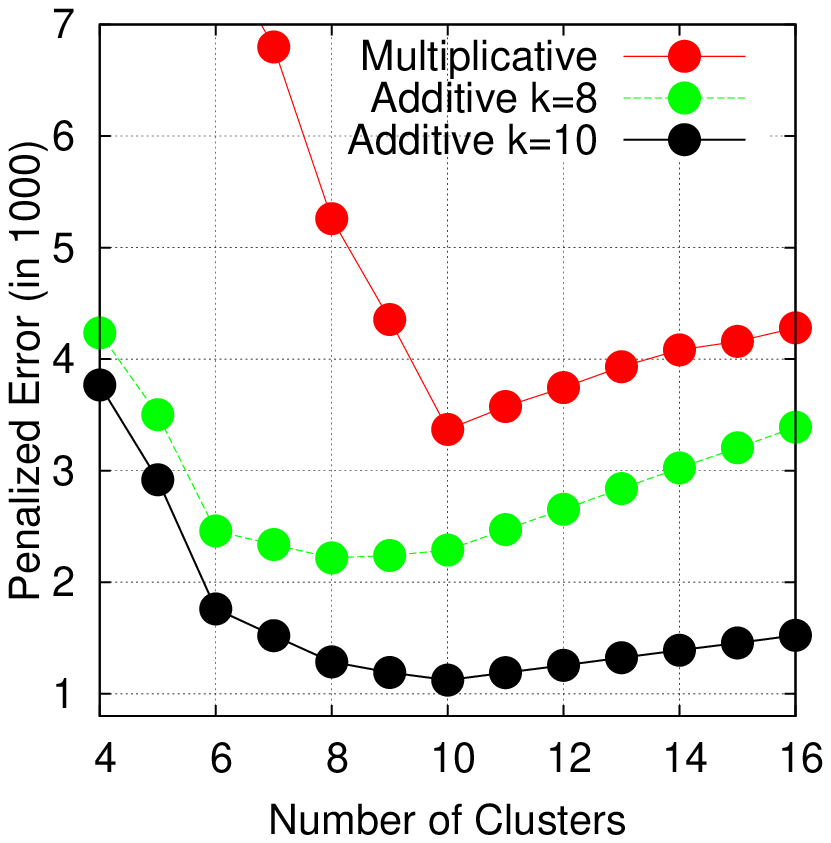}}
}
\caption{A 2D examples with 10 clusters, each cluster with
approximately 100 points. The figures in the top row show 
when $k\!=\!9$ one of the clusters is a dumbbell, and when
$k\!=\!11$ one of the clusters splits in half. The figure in the bottom row left
shows that the additive penalty yields $k\!=\! 3,6,8,10$ as solution candidates,
while the right figure shows that $k\!=\! 10$ is the only possible
solution.}
\end{figure}

\begin{figure}[th]
\center{
{\includegraphics[width=2.0in] {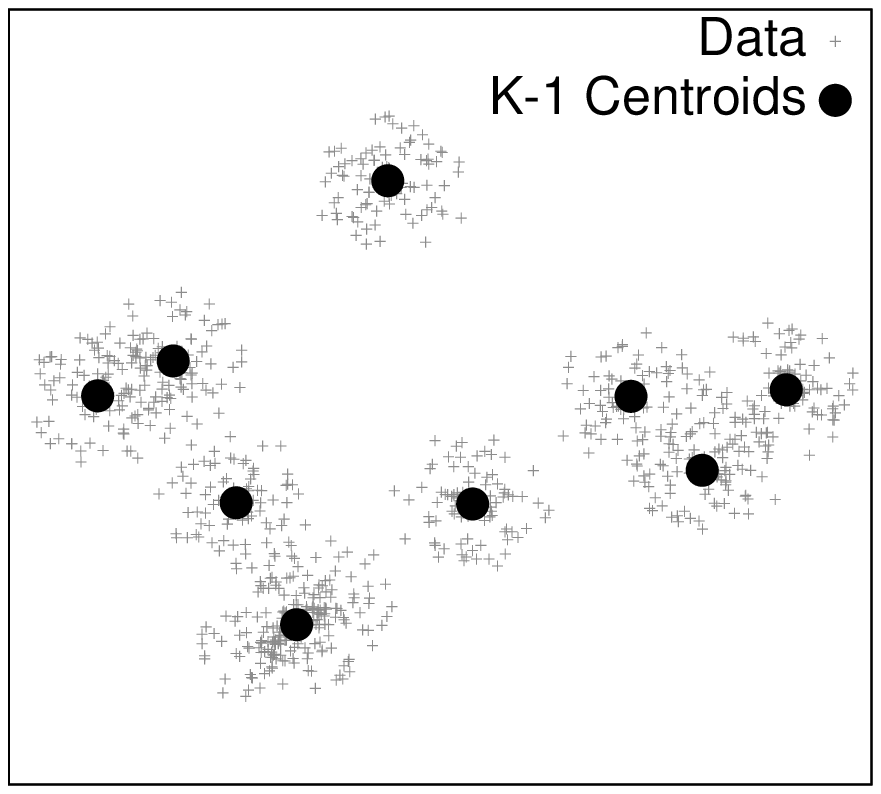}}
\hspace*{-0.5cm}
{\includegraphics[width=2.0in] {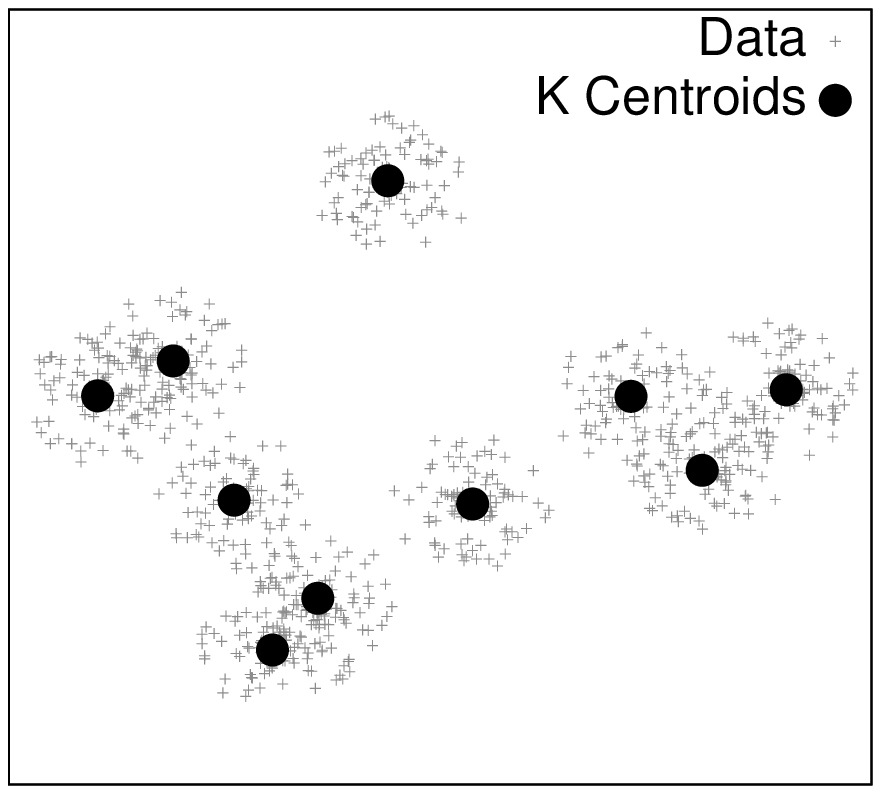}}
\hspace*{-0.5cm}
{\includegraphics[width=2.0in] {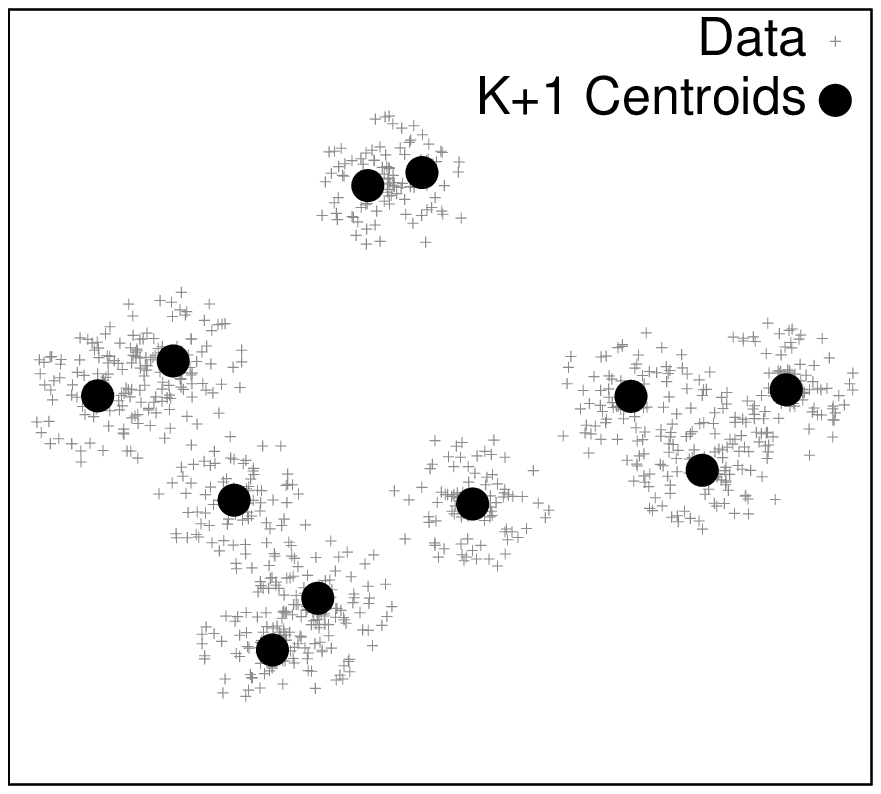}}\\
{\includegraphics[width=2.0in] {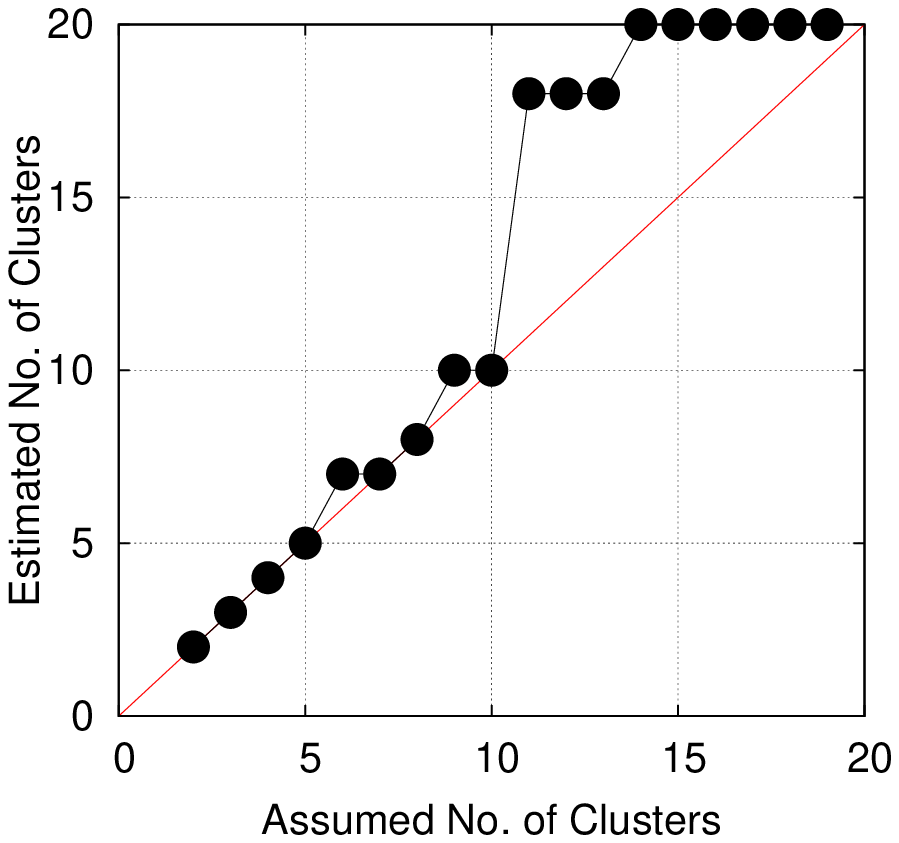}}
\hspace*{-0.6cm}
{\includegraphics[width=2.0in] {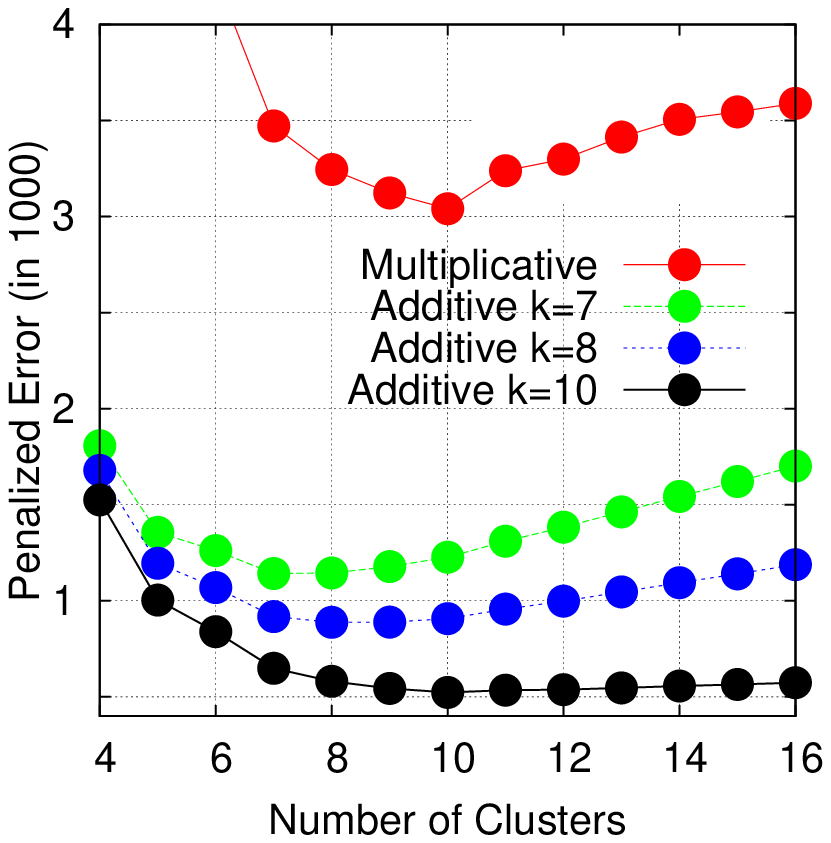}}
}
\caption{This example is similar to Fig. 2. Clusters are the same but their
locations are changed to create more overlaps, which means
they are further from the ideal cluster assumption than those in Fig. 2.
Additive penalty yields $k\!=\!2,3,4,5,7,8,10$ are solution candidates,
while the multiplicative penalty shows $k\!=\! 10$ is the possible solution.
}
\end{figure}

Fig. 4 is a 2D case with $K\!=\! 20$ clusters each with about 200 data points.
As can be seen there is little overlap among clusters, nevertheless additive
penalty yields several solutions while the multiplicative penalty yields
only one solution at the correct value. In Fig.~5 the clusters are the same as
in Fig.~4 but the inter-cluster distances have been significantly reduced so
that the clusters have substantial overlap. Even though both additive
and multiplicative penalties yield several solutions, they both only
agree on $K\!=\!16$.

\begin{figure}[t]
\center{
{\includegraphics[width=1.8in] {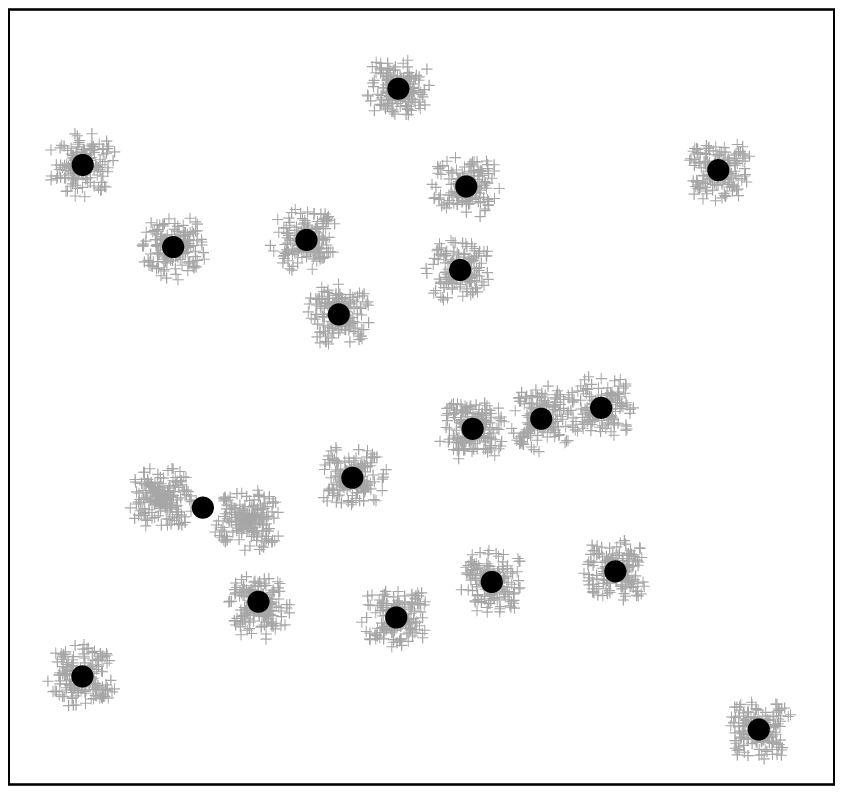}}
\hspace*{-0.5cm}
{\includegraphics[width=1.8in] {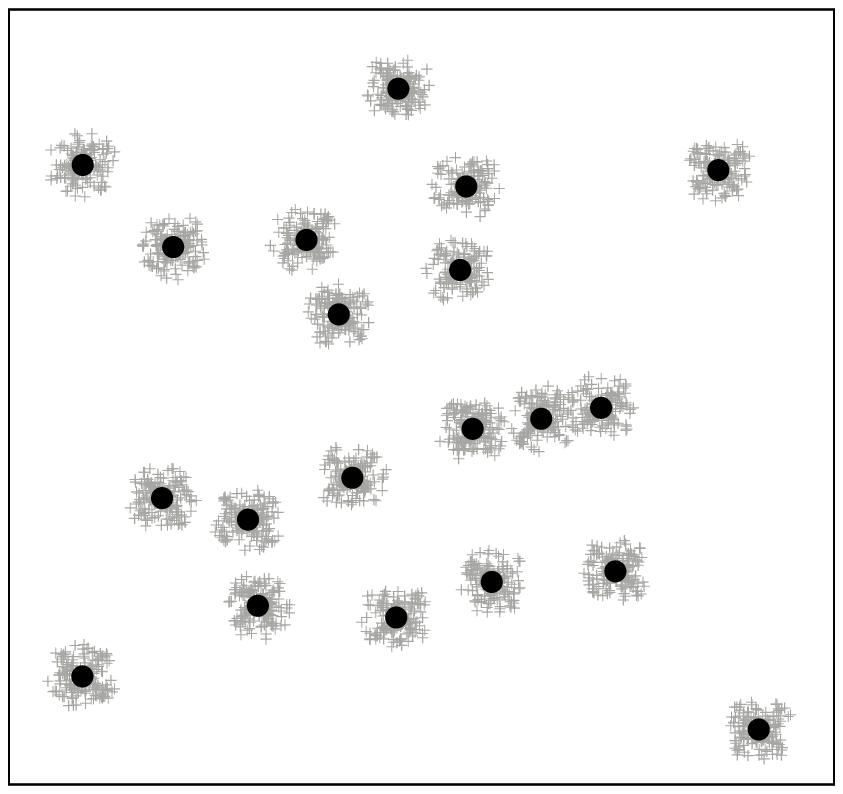}}
\hspace*{-0.5cm}
{\includegraphics[width=1.8in] {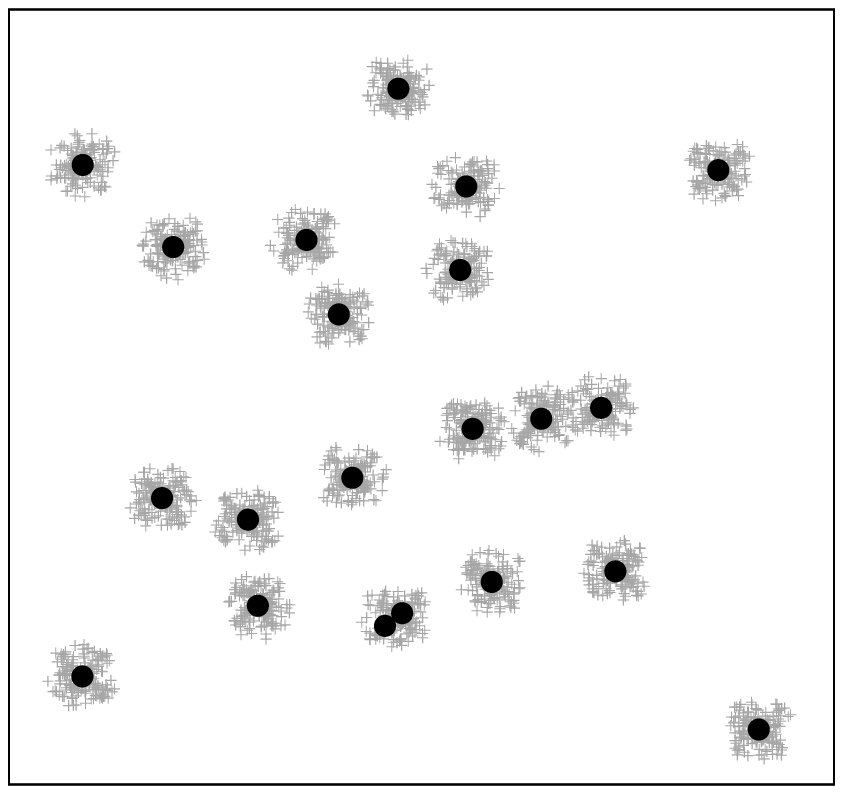}}
\hspace*{-0.5cm}
{\includegraphics[width=1.75in] {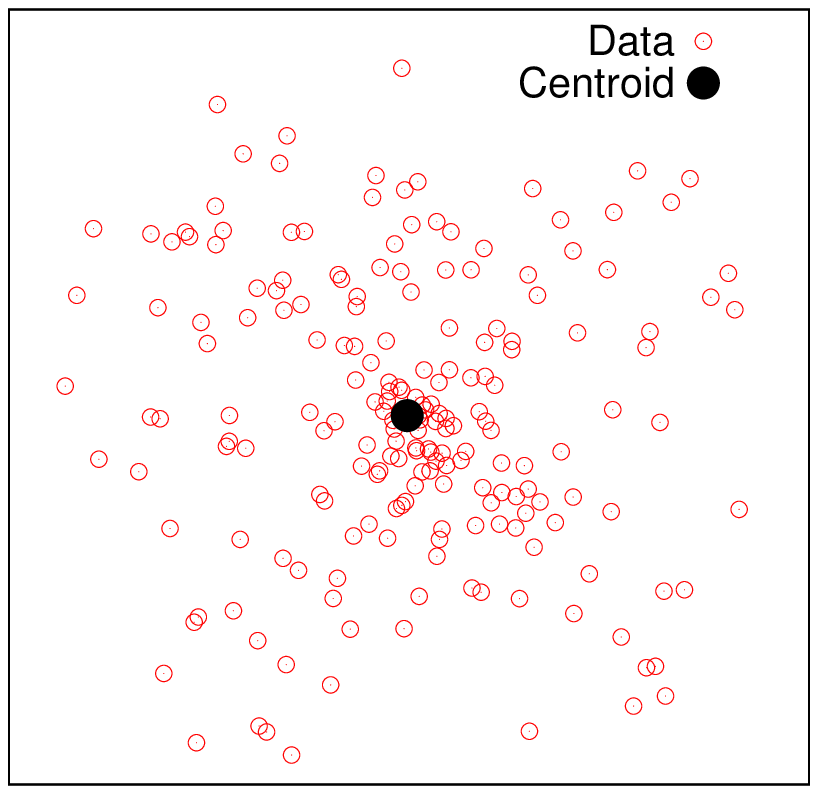}}\\
\vspace*{-0.1cm}
{\includegraphics[width=2.0in] {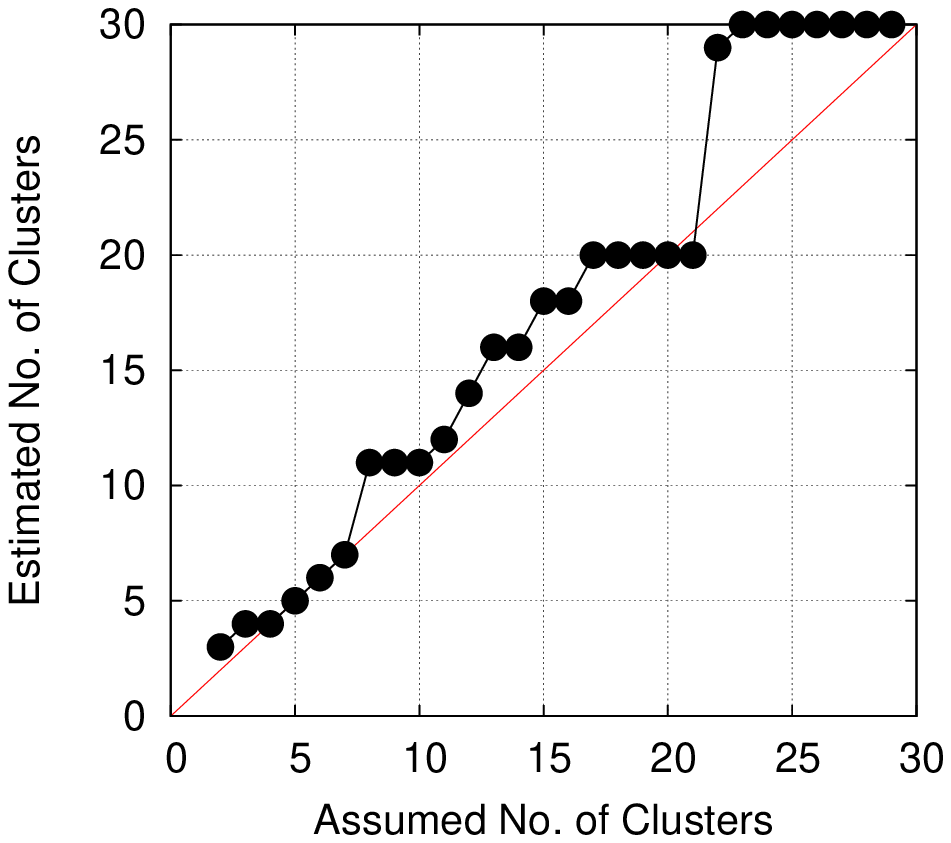}}
{\includegraphics[width=2.0in] {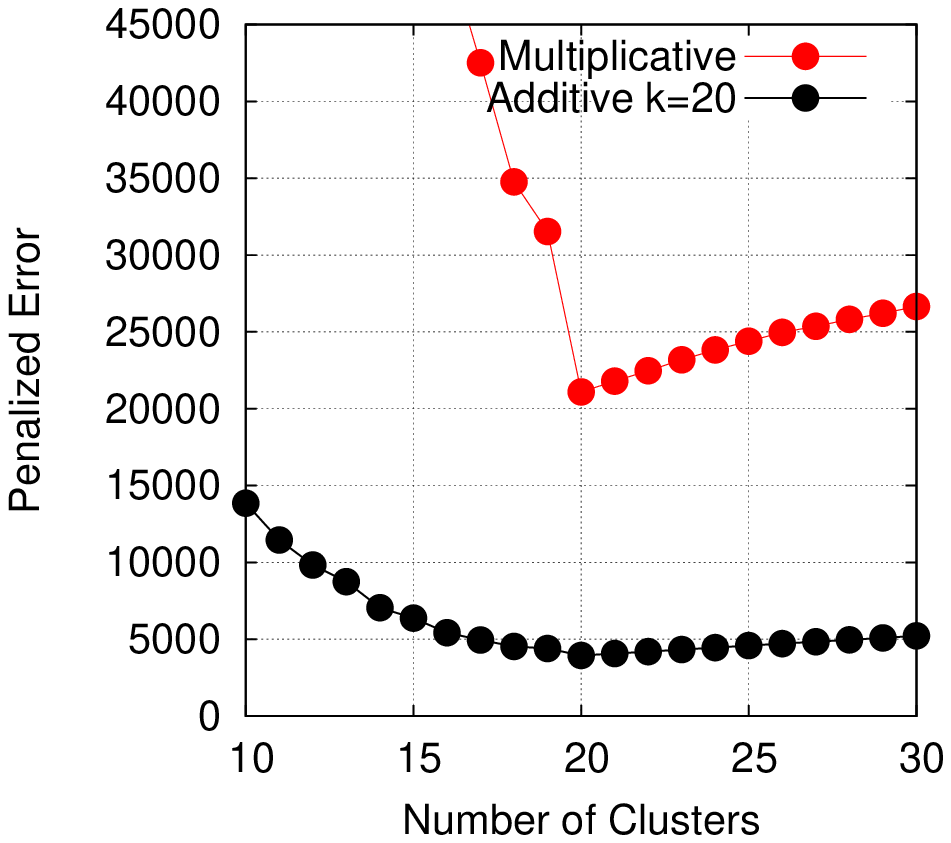}}
}
\caption{A 2D example with $K\!=\!20$, each cluster has approximately 200 
points. 
({\bf Row 1}) The three figures on the left show the result of clustering 
with $k\!=\!19,20,21$.
Note that when  $k\!=\!19$ one of the clusters is a dumbbell, and when
$k\!=\!21$ one of the clusters splits in half. Figure on the right shows
distribution of data points in a typical cluster.
({\bf Row 2}) Left figure shows {\em Estimated} vs. {\em Assumed} number of 
clusters for the additive penalty, with $K\!=\!4,5,6,7,20$ as candidate
solutions. Right figure shows that $E^{(m)}$ has a solution at
$K\!=\!20$ only. Also plotted is $E^{(a)}$ for $K\!=\!20$ which has a shallow 
minimum. 
}
\end{figure}

\begin{figure}[t]
\center{
{\includegraphics[width=2.0in] {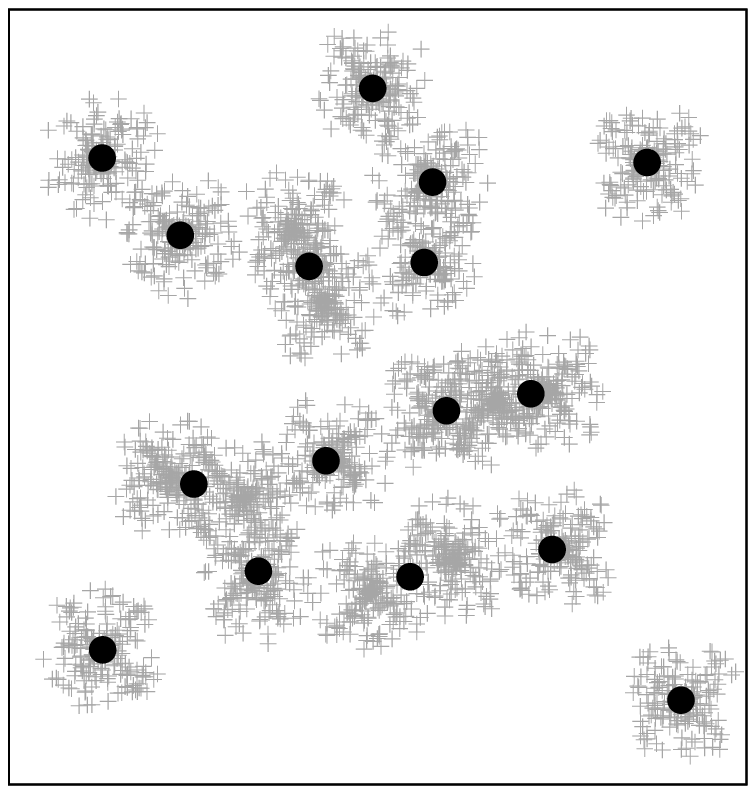}}
{\includegraphics[width=2.1in] {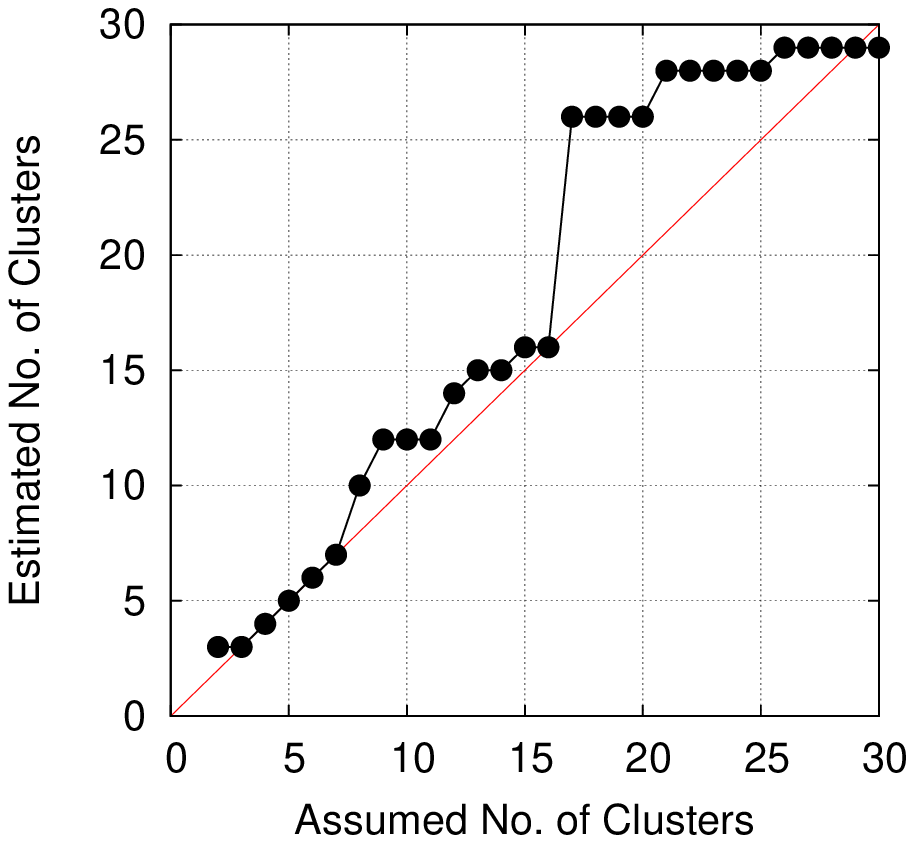}}
{\includegraphics[width=2.1in] {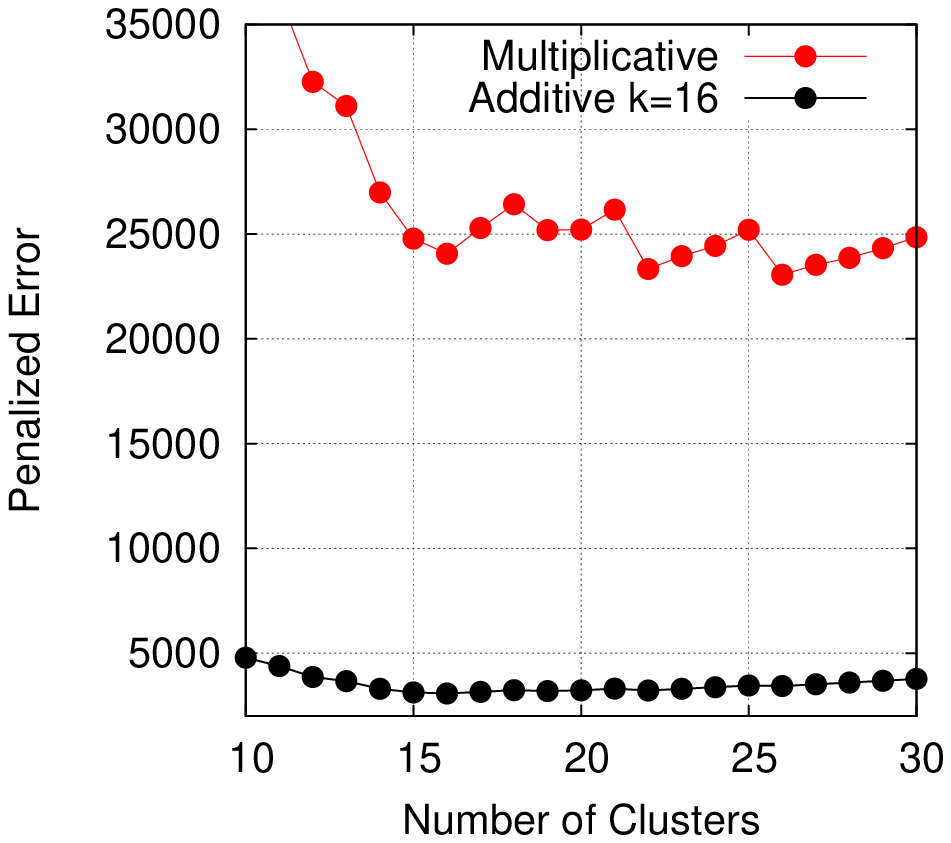}}
}
\caption{Clusters are the same as in Fig.~4, but the inter-cluster
distances are uniformly shrunk. The middle figure shows candidate
solutions for additive penalty
are $K\!=\!3,4,5,6,7,16,29$, while the figure on the right shows the 
local minima for multiplicative penalty are at
$K\!=\!16,19,22,26$. They agree only on $K\!=\!16$, a
reasonable solution shown on the left panel.   }
\end{figure}

Fig. 6 is an example with $K\!=\! 20$ clusters in $d\!=\! 8$ dimensions. 
The additive penalty yields $k\!=\! 2,20$ as candidate solutions, while the 
multiplicative penalty yields only  $k\!=\! 20$. We also plot $E^{(a)}$
for $K\!=\! 20$ and $E^{(m)}$, which show the comparative depth of their
minima. Note that at higher dimensions clusters generally have less 
overlap, thus becoming more like ideal clusters with little overlap
and unambiguous solutions.

Fig. 7 is a composite picture of five Brodatz textures with 
256$\times$256 image size. There are  many alternative ways of 
representing the image for clustering.
Here we convert the image to a set of data points into a $d\!=\!16$ dimensional
space by using non-overlapping 
4$\times$4 cosine transform. Thus, the image with over 65k pixels is 
represented by
by 4,096 data points in a 16-dimensional feature space. The pixels on the 
borders of the texture regions are sparse and appear like outliers. 
The clustering algorithm presented in Sec.~2.1 is sensitive to outliers, as
are other clustering algroithms. Therefore, to deal with outliers we
compute the point density around each of them, and eliminate those
that have densities below a certain threshold. The remaining data is
3,660 points.  In this figure we show the results of the penalized k-means
algorithms.
As may be seen, both additive and multiplicative penalties have
pronounced minima at $k\!=\!5$ 
yielding the correct number of clusters or textures. Again we note that 
clusters
in high dimensional spaces are more likely to behave as ideal clusters, hence
yielding unambiguous correct solutions.

\begin{figure}[t]
\center{
{\includegraphics[width=1.7in] {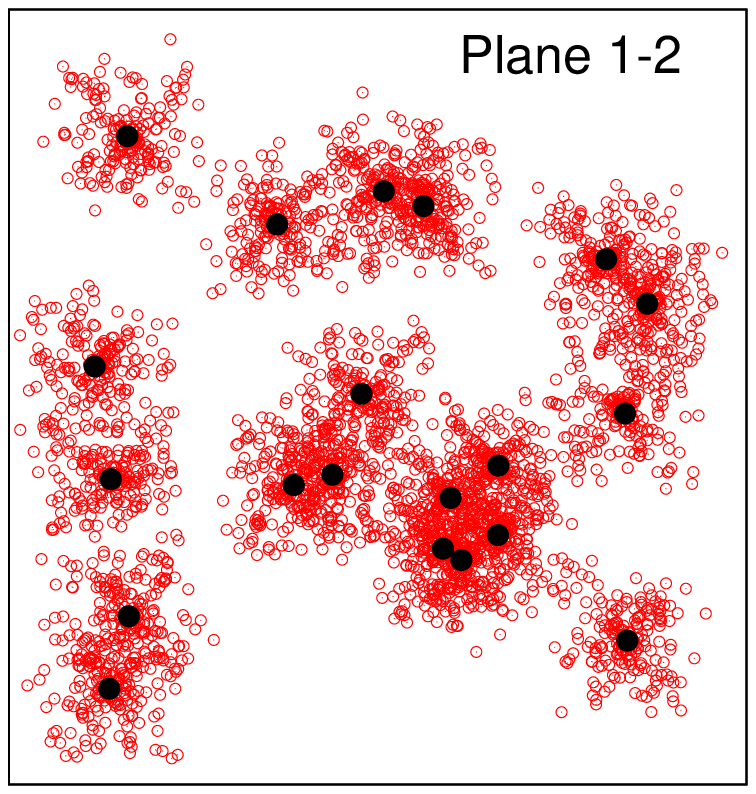}}
\hspace*{-0.6cm}
{\includegraphics[width=1.7in] {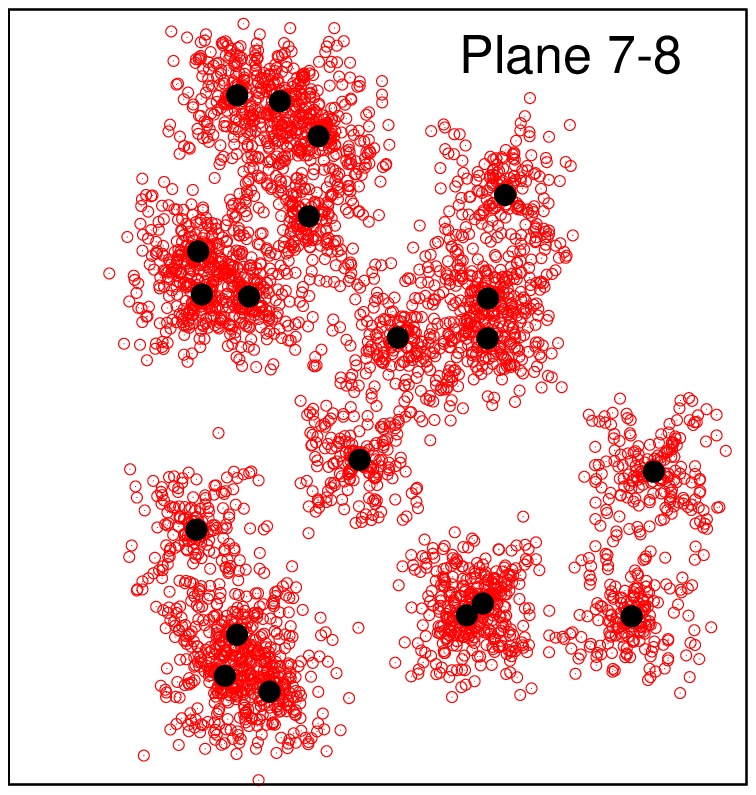}}
\hspace*{-0.4cm}
{\includegraphics[width=1.9in] {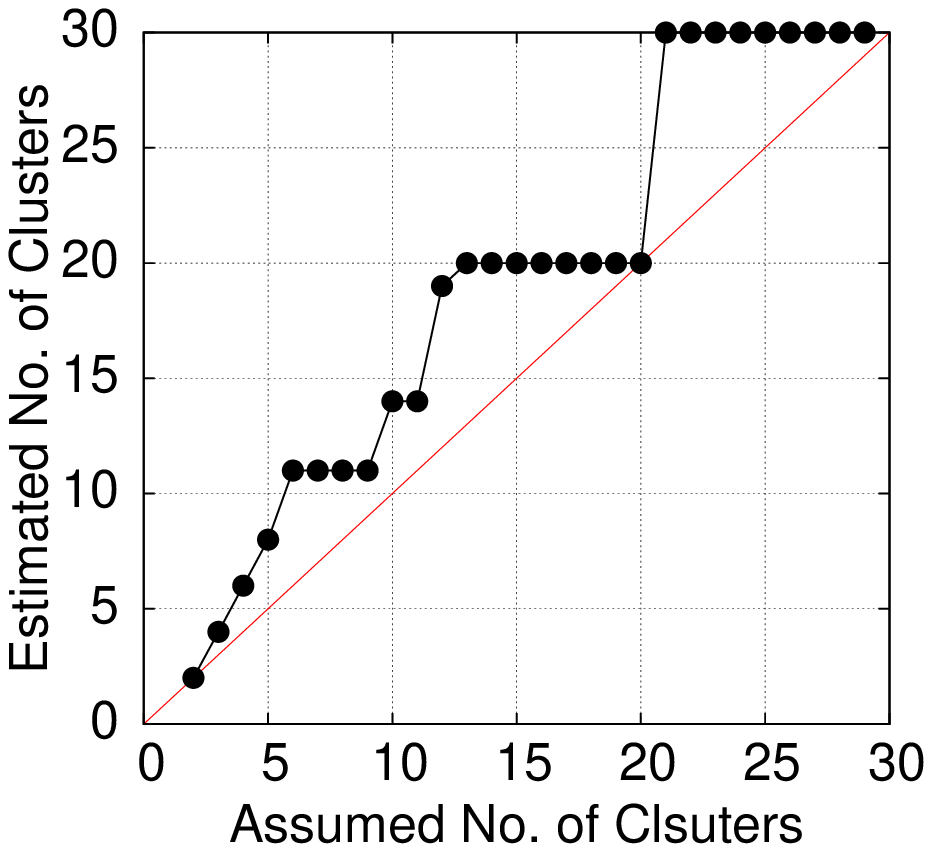}}
\hspace*{-0.8cm}
{\includegraphics[width=1.9in] {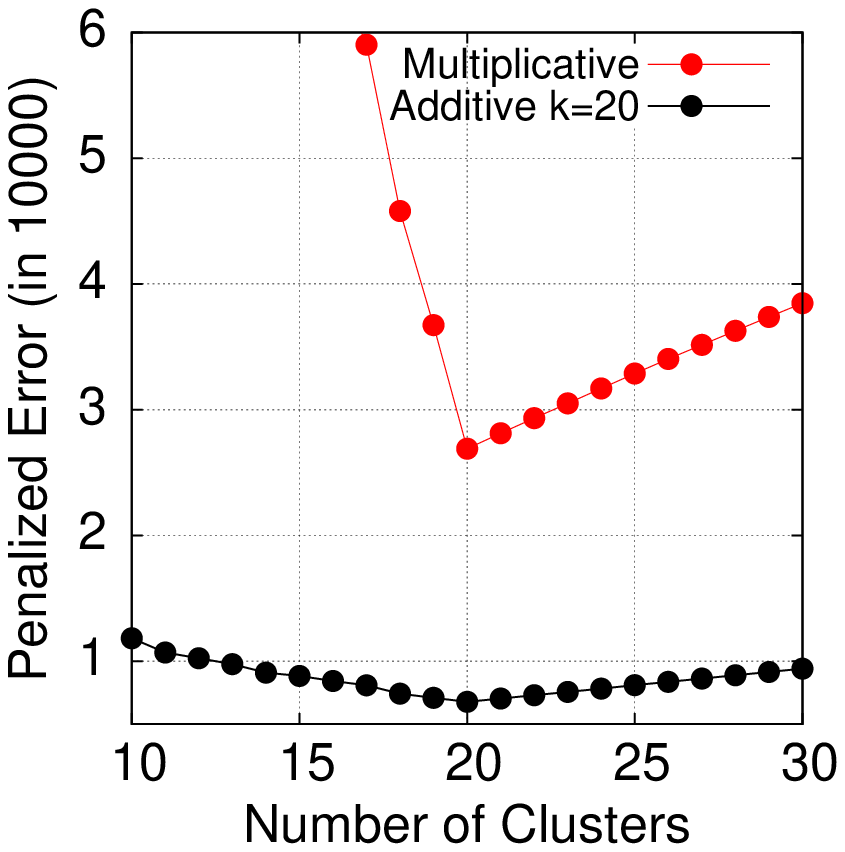}}
}
\caption{An example with 20 clusters in 8 dimensions. Right two figures
are projections of clusters on two different 2D planes and create the
impression they have significant overlap. 
The actual clusters are sufficiently
well separated, hence penalized k-means yields unambiguous solutions.}
\end{figure}

\begin{figure}[t]
\center{
{\includegraphics[width=1.5in] {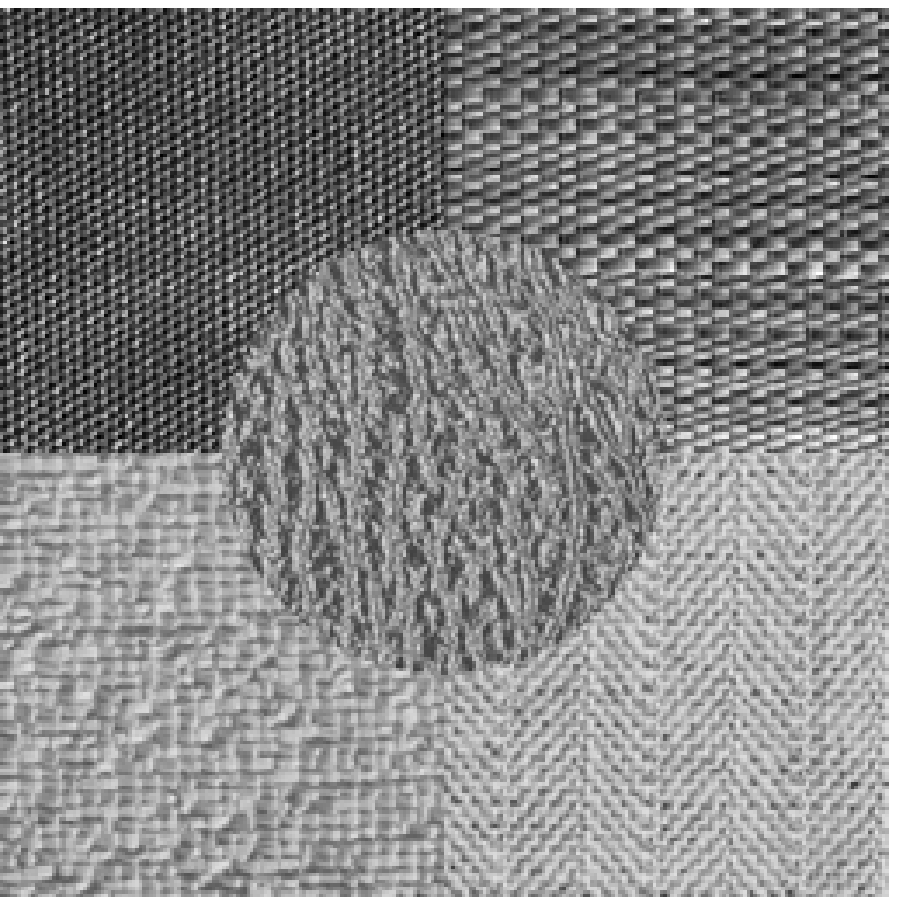}}
\hspace*{0.3cm}
{\includegraphics[width=1.8in] {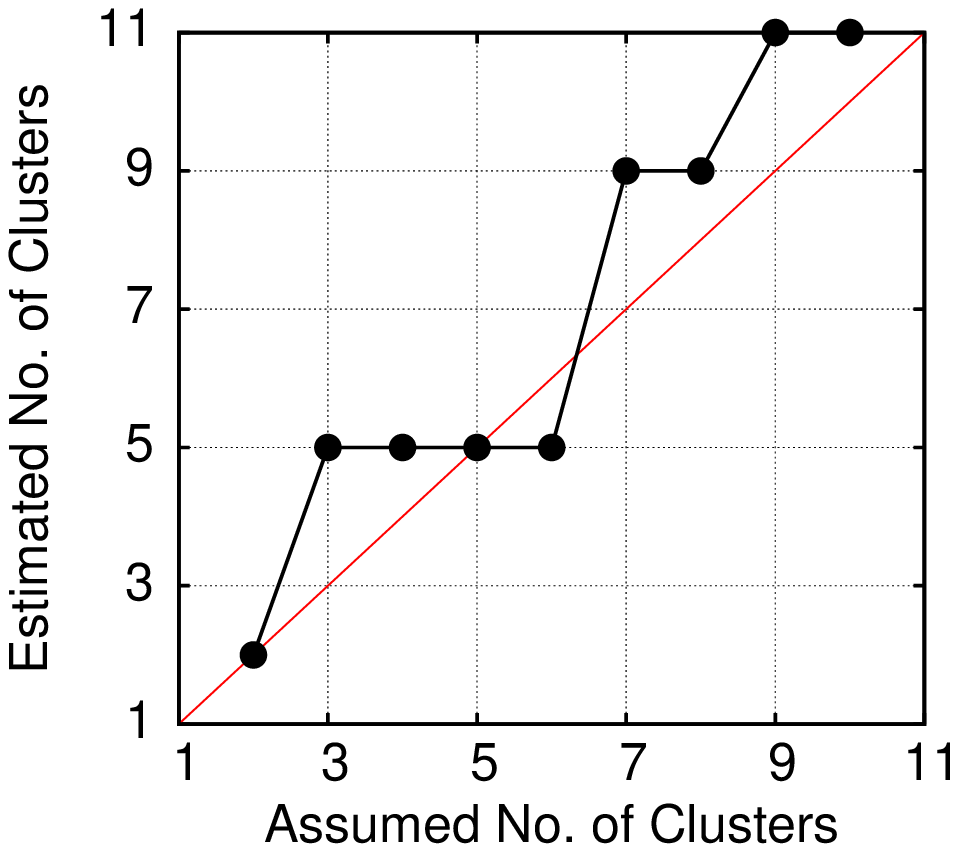}}
{\includegraphics[width=1.8in] {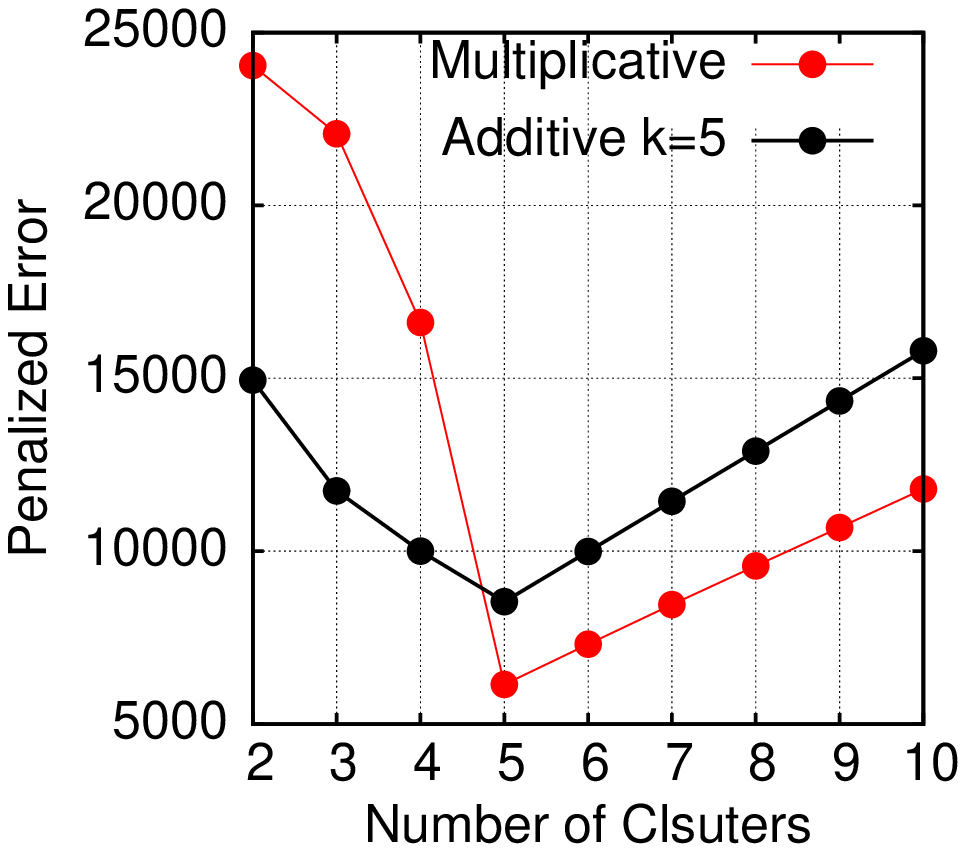}}
}
\caption{The picture is a composite of 5 Brodatz textures. The image is 
transformed 
into a $d\!=\! 16$ dimensional space. The k-means algorithms with both
additive and multiplicative penalties yield 5 clusters (textures).  }
\end{figure}

\section{Concluding Remarks}

Penalized k-means with additive penalty is a popular algorithm
for finding the correct number of clusters in a dataset. 
However, in practice the coefficient of the additive penalty term 
is chosen arbitrarily in an ad hoc manner. 
We presented a principled derivation of the bounds for the coefficient of 
additive penalty in k-means for ideal clusters. 
Even though in practice clusters
typically deviate from the ideal assumption, the  ideal case
serves as a useful guideline. We also investigated multiplicative penalty, 
which turns out to produce a more reliable signature for the correct number
of clusters. We examined empirically deviations from the {\em non-overlap} 
assumption and presented a procedure
for combining k-means with additive and multiplicative penalties to obtain the
correct number of clusters when one or both approaches yield ambiguous 
solutions. What the limits of this disambiguation procedure are
needs further theoretical analysis or more exhaustive empirical evidence.

This work may be further extended in a number of directions that
relax the ideal cluster assumptions. These include (a) 
rigorous derivation of the limits of non-overlap assumption;
(b) extend
to spherically symmetric normal distribution of data points rather than
spherical distributions, which would allow
greater tolerance of overlapping clusters; (c) relax the assumption that
the number of points in each cluster be approximately the same; (d) 
theoretical analysis and the algorithm for finding the best clustering is 
sensitive to outliers, develop more sophisticated methods for detecting
and discarding outliers.


\pagebreak

\appendix
\section{Appendix A}

In this Appendix, we calculate the volume of a sphere in the $n$-dimensional
Euclidean space, as well as the within-cluster variances for the
sphere, and the centroid and the within-cluster variances
for the half-sphere and a dumbbell. Note that while in the body of the paper we
denote the dimension by $d$, in this appendix we denote it by
$n$ to avoid confusion with the differential element $d$ in the integrals.

\subsection*{A.1 Sphere}

The volume element in the $n$-dimensional spherical coordinate system
 (in terms of radius $r$ and angles $\phi_1,\ldots, \phi_{n-1}$) is:
\begin{equation}
dV=
[dr\,r^{n-1}]\,[d\phi_1\,(\sin{\phi_1})^{n-2}]\,
[d\phi_2\,(\sin{\phi_2})^{n-3}]\,\cdots\,
[d\phi_{n-2}\,(\sin{\phi_{n-2}})]\,
[d\phi_{n-1}].
\end{equation}
Hence the volume of a sphere of radius $R$ may be calculated from:
\begin{equation}
\begin{array}{ll}
V=&
\displaystyle[\int_{0}^{R}\,dr\,r^{n-1}]\,[2\int_{0}^{\pi/2}d\phi_1\,(\sin{\phi_1})^{n-2}]\,
[2\int_{0}^{\pi/2}d\phi_2\,(\sin{\phi_2})^{n-3}]\,\cdots\\
\, &\, \\
 & \displaystyle[2\int_{0}^{\pi/2}d\phi_{n-2}\,(\sin{\phi_{n-2}})]\,
[4\int_{0}^{\pi/2}d\phi_{n-1}].
\end{array}
\end{equation}
The integrals over spherical angles $\phi_1 \cdots \phi_{n-1}$ can be
calculated from the following formula~\cite{Grad},
\begin{equation}
\int_{0}^{\pi/2}d\phi\,(\sin{\phi})^{n}=
\frac{\Gamma(\displaystyle\frac{n+1}{2})\Gamma(\displaystyle\frac{1}{2})}{2\,\Gamma(\displaystyle\frac{n+2}{2})},
\,\,\,\,\,\mbox{for all integers}\,\,\,n\ge 0,
\end{equation}
where $\Gamma(z\!+\!1)\!=\!z\Gamma(z)$, the generalized factorial function, is:
\begin{equation}
\mbox{For integer}\,\, n,\,\,\,\,
\Gamma(n)=(n-1)!\,\,\,\,\mbox{and}\,\,\,\,\, 
\Gamma(n+\frac{1}{2})=\frac{(2n)!\sqrt{\pi}}{4^n\,n!}.
\end{equation}
With repeated use of (A.17), Eq. (A.16) simplifies to:
\begin{equation}
V=\displaystyle\frac{\pi^{\frac{n}{2}}}{\Gamma(\displaystyle\frac{n+2}{2})}\,R^n.
\end{equation}

\begin{figure}[t]
\center{
{\includegraphics[width=3.0in] {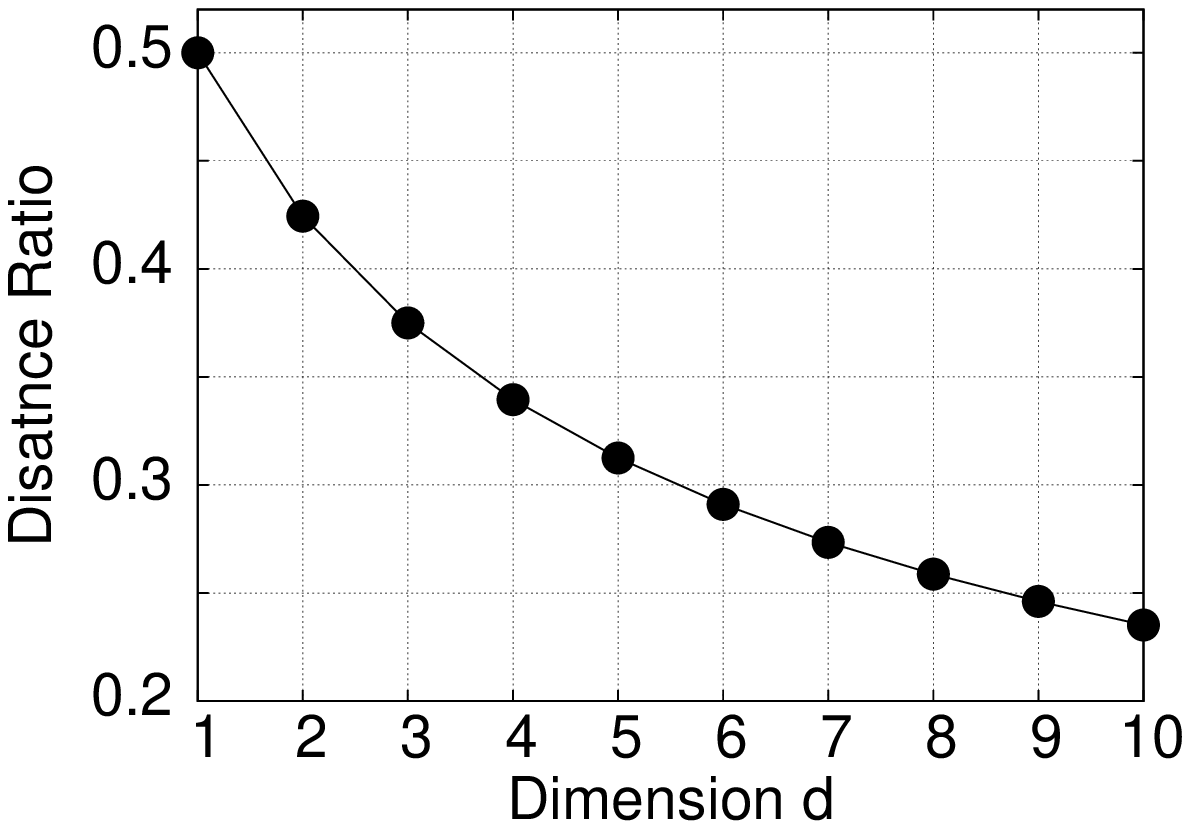}}
{\includegraphics[width=3.0in] {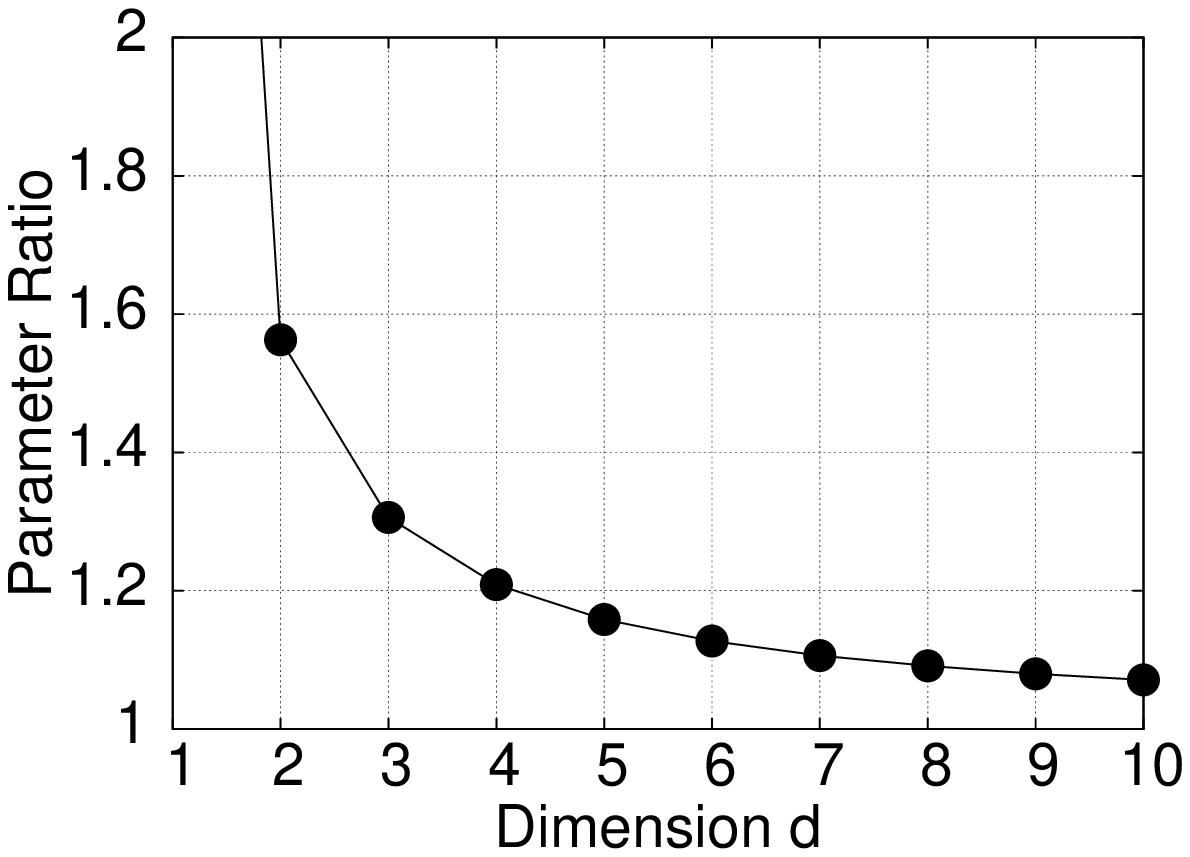}}
}
\caption{({\bf Left}) The ratio $\gamma=\rho/R$ as a function of dimension $d$,
where $R$ is the sphere radius and $\rho$ is distance of the half-sphere
centroid from its equator plane;
$\rho\!\rightarrow\! 0$ as $d\!\rightarrow\!\infty$.
({\bf Right}) The ratio $\alpha/2\beta$ as a function of dimension $d$.
The ratio goes to 1 as $d\!\rightarrow\!\infty$.}
\end{figure}

The within-cluster variance, or clustering error, for the $n$-dimensional
sphere is given by:
\begin{equation}
E_s=\int_{sphere}\,dV \parallel \mathbf{x}-\mathbf{c}\parallel^2,
\end{equation}
 and may be expressed in the spherical coordinate system (the centroid
 $\mathbf{c}$ is at the origin),
\begin{equation}
\begin{array}{ll}
E_s=&
\displaystyle[\int_{0}^{R}\,dr\,r^{n+1}]\,[2\int_{0}^{\pi/2}d\phi_1\,(\sin{\phi_1})^{n-2}]\,
[2\int_{0}^{\pi/2}d\phi_2\,(\sin{\phi_2})^{n-3}]\,\cdots\\
\, &\, \\
 & \displaystyle[2\int_{0}^{\pi/2}d\phi_{n-2}\,(\sin{\phi_{n-2}})]\,
[4\int_{0}^{\pi/2}d\phi_{n-1}].
\end{array}
\end{equation}
This can be similarly simplified to:
\begin{equation}
E_s=V\,R^2\,\alpha,\,\,\,\,\, \mbox{with}\,\,\,
\alpha=\frac{n}{n+2}.
\end{equation}

\subsection*{A.2 Half-sphere}

The centroid $\mathbf{c}$ of the half-sphere is on the
axis that is orthogonal to the equator, say axis 1, that is
$\mathbf{c}=(\rho,0,0,\ldots,0)$. Its distance, $\rho$, from the equator
may be calculated from
\begin{equation}
\begin{array}{ll}
\frac{1}{2}\,V\,\rho&= \displaystyle\int_{half-sphere} dV \,r\\
\, & \,\\
 \,&=\displaystyle[\int_{0}^{R}\,dr\,r^{n}]\,[\int_{0}^{\pi/2}d\phi_1\,(\sin{\phi_1})^{n-2}
\cos{\phi_1}]\,
[2\int_{0}^{\pi/2}d\phi_2\,(\sin{\phi_2})^{n-3}]\,\cdots\\
\, &\, \\
 &\,\,\,\,\,\,\displaystyle[2\int_{0}^{\pi/2}d\phi_{n-2}\,(\sin{\phi_{n-2}})]\,
[4\int_{0}^{\pi/2}d\phi_{n-1}].
\end{array}
\end{equation}
Subscript $h$ indicates integral over the volume of the half-sphere.
This leads to
\begin{equation}
\rho=R\gamma, \,\,\,\,\,\mbox{with}\,\,\,\,\gamma=\frac{\Gamma(\displaystyle\frac{n+2}{2})}{\sqrt{\pi}\,\Gamma(\displaystyle\frac{n+3}{2})}.
\end{equation}

The within-cluster variance for the half-sphere is given by:
\begin{equation}
E_h=\int_{half-sphere}
\,dV\,\parallel\mathbf{r}-\mathbf{c}\parallel^2,
\end{equation}
which reduces to the following,
\begin{equation}
E_h=\frac{1}{2}\,E_s-\frac{1}{2}\,V\,\rho^{2}=V\,R^2\,\beta,\,\,\,\,\,
\mbox{with}\,\,\,
\beta=\frac{1}{2}\,(\frac{n}{n+2} - \gamma^2)=\frac{1}{2}(\alpha-\gamma^2).
\end{equation}

\subsection*{A.3 Dumbbell}

The within-cluster variance for the dumbbell with two equal size
spheres may be expressed as,
\begin{equation}
E_d= 2E_s+ 2VL^2 =2 VR^2\alpha +2VL^2,
\end{equation}
where $L$ is the distance of the center of the dumbbell to the center of either
sphere.

A quantity of interest is the ratio $\displaystyle\frac{\alpha}{2\beta}$,
which ranges between 1 and 4.  For $n\!=\! 1$ (straight line) we have
$\displaystyle\frac{\alpha}{2\beta}\!=\! 4$, and as $n\rightarrow \infty$
it approaches 1,
thus $1\!<\!\displaystyle\frac{\alpha}{2\beta} \!\le\! 4$.  See Fig.~8.

\section{Appendix B}

In this Appendix we derive bounds for the coefficient $\lambda$ in the
additive penalized
k-means when the penalty function has polynomial, logarithmic, and exponential
dependence on the number of clusters $K$.
For the additive penalty, Eq. (1.2), in order to get a similar signature to
multiplicative penalty case, we must require the following inequalities:
\begin{equation}
\left\{
\begin{array}{lll}
\Delta_{K-1,K}^{(a)}\!\!&=E_{K-1}-E_{K}=2VL^2-
\lambda(f(K)-f(K-1))>0& \mbox{for all}\,\,\,n\ge 1
\,\,\,\mbox{and}\,\,\,K\ge 1,\\
 & & \\
\Delta_{K,K+1}^{(a)}\!\!&=E_K-E_{K+1}=VR^2(\alpha-2\beta)-
\lambda(f(K+1)-f(K))<0&
\mbox{for all}\,\,\,n\ge 1 \,\,\,\mbox{and}\,\,\,K\ge 1.\\
\end{array}
\right.
\end{equation}
Noting that $R^2(\alpha-2\beta)\!=\!\rho^2$, the two inequalities in (B.1)
maybe be combined and written as:
\begin{equation}
\frac{V\rho^2}{f(K+1)-f(K)}<\lambda<\frac{2VL^2}{f(K)-f(K-1)}.
\end{equation}
To estimate the value of $\lambda$, we remember that the volume $V$ is
the average number of data points in each cluster, i.e., $V\approx N/K$,
$\rho$ is given by (A.24) and is less than $R/2$ and goes to zero as the
dimension $n$ increases. And $L$ is an  inter-cluster distance.

We consider three functional forms for $f(K)=\ln K, K^p, e^K$. For these
penalty terms (B.29) becomes,
\begin{equation}
\left\{
\begin{array}{ll}
f(K)=\ln K: & \displaystyle{N\rho^{2}[1+\frac{1}{K}+O(\frac{1}{K^2})] < \lambda 
< 2NL^2[1+\frac{1}{K}+O(\frac{1}{K^2})]}\\
 & \\
f(K)=K^p: & 
\displaystyle{\frac{N\rho^{2}}{pK^{p}}[1-\frac{p-1}{2K}+O(\frac{p^2}{K^2})]<
\,\lambda  \, 
     < \frac{2NL^2}{pK^{p}}[1+\frac{p-1}{2K}+O(\frac{p^2}{K^2})]}\\
 & \\
f(K)=e^K: & \displaystyle{\frac{N\rho^2}{(e-1)e^K} < \lambda  < 
\frac{2NL^2}{(1-\frac{1}{e})e^K}},
\end{array}
\right.
\end{equation}
where we assume $K$ is sufficiently greater than $p$ for the polynomial case.
It can be seen that $\lambda$ depends on the correct number of clusters $K$,
which is not known a priori. In particular, for the commonly used linear 
penalty function, $f(K)=K$, we get:
\begin{equation}
\frac{N\rho^2}{K} < \lambda  < \frac{2NL^2}{K}.
\end{equation}

\pagebreak

\end{document}